\documentclass[10pt,twocolumn,letterpaper]{article}

\usepackage{epigraph} 
\usepackage{cvpr}              
\usepackage{url} 
\usepackage[accsupp]{axessibility}
\usepackage{cuted}
\usepackage{capt-of}
\usepackage{times}
\usepackage{epsfig}
\usepackage{graphicx}
\usepackage{enumitem}
\usepackage{amsmath}
\usepackage{amssymb}
\usepackage{xcolor} 

\usepackage{multirow}

\usepackage[square,sort,comma,numbers]{natbib}

\newcommand{\figdir}{figures}

\definecolor{citecolor}{HTML}{0071BC}
\definecolor{linkcolor}{HTML}{ED1C24}
\usepackage[pagebackref=false,breaklinks=true,letterpaper=true,colorlinks,citecolor=citecolor,linkcolor=linkcolor,bookmarks=false]{hyperref}

\usepackage[capitalize]{cleveref}
\crefname{section}{Sec.}{Secs.}
\Crefname{section}{Section}{Sections}
\Crefname{table}{Table}{Tables}
\crefname{table}{Tab.}{Tabs.}

\def\cvprPaperID{6522} 
\def\confName{CVPR}
\def\confYear{2023}

\begin{document}

\title{Delving StyleGAN Inversion for Image Editing:\\ A Foundation Latent Space Viewpoint}


\author{
    \vspace{0.4em}
    \centerline{
  Hongyu Liu$^{1}$\qquad
  Yibing Song$^{2}$\thanks{Y. Song and Q. Chen are the joint corresponding authors.}\qquad
  Qifeng Chen$^{1*}$} 
  \\
  $^{1}$Hong Kong University of Science and Technology \quad 
  $^{2}$AI$^3$ Institute, Fudan University\\
  \tt\small{hliudq@cse.ust.hk \qquad yibingsong.cv@gmail.com \qquad cqf@ust.hk} \\
}
\maketitle

\begin{strip}
\vspace{-0.7in}
\setlength\tabcolsep{0.5pt}
\centering
\includegraphics[width=1.0\linewidth]{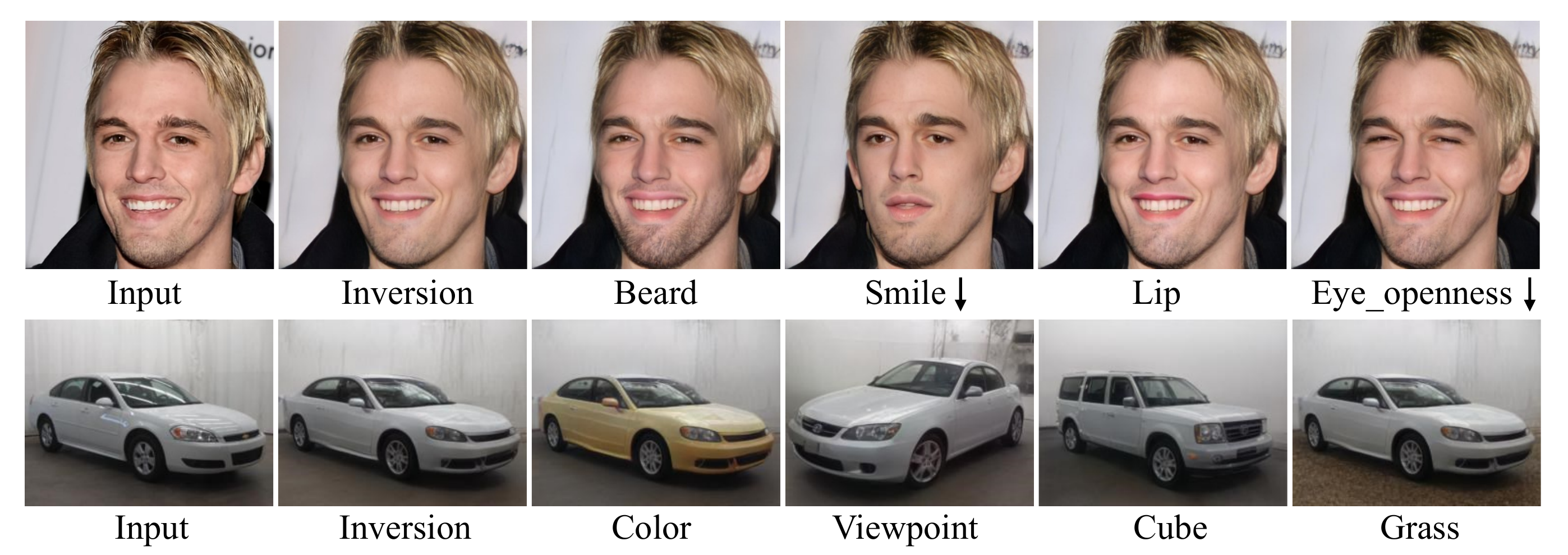}
\vspace{-8pt}
\captionof{figure}{The inversion and editing results of our model in the real images. We show from the left to right of each row: an input image, inversion results, and our editing results. We edit images by modifying the attributes in the embedding space following~\cite{shen2020interpreting,harkonen2020ganspace}. The $\downarrow$ means a decreased magnitude of the manipulation attribute.}
\vspace{-8pt}
\label{fig:teaser}
\end{strip}

\begin{abstract}

GAN inversion and editing via StyleGAN maps an input image into the embedding spaces ($\mathcal{W}$, $\mathcal{W^+}$, and $\mathcal{F}$) to simultaneously maintain image fidelity and meaningful manipulation. From latent space $\mathcal{W}$ to extended latent space $\mathcal{W^+}$ to feature space $\mathcal{F}$ in StyleGAN, the editability of GAN inversion decreases while its reconstruction quality increases. Recent GAN inversion methods typically explore $\mathcal{W^+}$ and $\mathcal{F}$ rather than $\mathcal{W}$ to improve reconstruction fidelity while maintaining editability. As $\mathcal{W^+}$ and $\mathcal{F}$ are derived from $\mathcal{W}$ that is essentially the foundation latent space of StyleGAN, these GAN inversion methods focusing on $\mathcal{W^+}$ and $\mathcal{F}$ spaces could be improved by stepping back to $\mathcal{W}$. In this work, we propose to first obtain the proper latent code in foundation latent space $\mathcal{W}$. We introduce contrastive learning to align $\mathcal{W}$ and the image space for proper latent code discovery. 
Then, we leverage a cross-attention encoder to transform the obtained latent code in $\mathcal{W}$ into $\mathcal{W^+}$ and $\mathcal{F}$, accordingly. Our experiments show that our exploration of the foundation latent space $\mathcal{W}$ improves the representation ability of latent codes in $\mathcal{W^+}$ and features in $\mathcal{F}$, which yields state-of-the-art reconstruction fidelity and editability results on the standard benchmarks. Project page: \url{https://kumapowerliu.github.io/CLCAE}.

\end{abstract}

\section{Introduction}
StyleGAN~\cite{karras2020analyzing,karras2019style,karras2020training} achieves numerous successes in image generation. Its semantically disentangled latent space enables attribute-based image editing where image content is modified based on the semantic attributes. GAN inversion~\cite{xia2021gan} projects an input image into the latent space, which benefits a series of real image editing methods~\cite{alaluf2021matter,patashnik2021styleclip,lin2021anycost,yuksel2021latentclr,zhu2022one}. The crucial part of GAN inversion is to find the inversion space to avoid distortion while enabling editability. Prevalent inversion spaces include the latent space $\mathcal{W^+}$~\cite{abdal2019image2stylegan} and the feature space $\mathcal{F}$~\cite{kang2021gan}. $\mathcal{W^+}$ is shown to balance distortion and editability~\cite{tov2021designing,zhu2020improved}. It attracts many editing methods~\cite{abdal2019image2stylegan,abdal2020image2stylegan++,guan2020collaborative,roich2021pivotal,alaluf2021restyle,hu2022style} to map real images into this latent space. On the other hand, $\mathcal{F}$ contains spatial image representation and receives extensive studies from the image embedding~\cite{kang2021gan,parmar2022spatially,xuyao2022,wang2022high} or  StyleGAN's parameters~\cite{alaluf2022hyperstyle,dinh2022hyperinverter} perspectives.

The latent space $\mathcal{W^+}$ and feature space $\mathcal{F}$ receive wide investigations. In contrast, Karras et al.~\cite{karras2020analyzing} put into exploring $\mathcal{W}$ and the results are unsatisfying. This may be because that manipulation in $\mathcal{W}$ will easily bring content distortions during reconstruction~\cite{tov2021designing}, even though $\mathcal{W}$ is effective for editability. Nevertheless, we observe that $\mathcal{W^+}$ and $\mathcal{F}$ are indeed developed from $\mathcal{W}$, which is the foundation latent space in StyleGAN. In order to improve image editability while maintaining reconstruction fidelity (i.e., $\mathcal{W^+}$ and $\mathcal{F}$), exploring $\mathcal{W}$ is necessary. Our motivation is similar to the following quotation:


\textit{``You can't build a great building on a weak foundation. You must have a solid foundation if you're going to have a strong superstructure.''}

\rightline{\textit{---Gordon B. Hinckley}}
 
In this paper, we propose a two-step design to improve the representation ability of the latent code in $\mathcal{W^+}$ and $\mathcal{F}$. First, we obtain the proper latent code in $\mathcal{W}$. Then, we use the latent code in $\mathcal{W}$ to guide the latent code in $\mathcal{W^+}$ and $\mathcal{F}$. In the first step, we propose a contrastive learning paradigm to align the  $\mathcal{W}$ and image space. This paradigm is derived from CLIP~\cite{radford2021learning} where we switch the text branch with $\mathcal{W}$. Specifically, we construct the paired data that consists of one image $I$ and its latent code $w\in\mathcal{W}$ with pre-trained StyleGAN. During contrastive learning, we train two encoders to obtain two feature representations of $I$ and $w$, respectively. These two features are aligned after the training process. During GAN inversion, we fix this contrastive learning module and regard it as a loss function. This loss function is set to make the one real image and its latent code $w$ sufficiently close. This design improves existing studies~\cite{karras2020analyzing} on $\mathcal{W}$ that their loss functions are set on the image space (i.e., similarity measurement between an input image and its reconstructed image) rather than the unified image and latent space. The supervision on the image space only does not enforce well alignment between the input image and its latent code in $\mathcal{W}$.



After discovering the proper latent code in $\mathcal{W}$, we leverage a cross-attention encoder to transform $w$ into $w^+\in\mathcal{W^+}$ and $f\in\mathcal{F}$. When computing $w^+$, we set $w$ as the query and $w^+$ as the value and key. Then, we calculate the cross-attention map to reconstruct $w^+$. This cross-attention map enforces the value $w^+$ close to the query $w$, which enables the editability of $w^+$ to be similar to that of $w$. Besides, $w^+$ is effective in preserving the reconstruction ability. When computing $f$, we set the $w$ as the value and key, while setting $f$ as the query. So $w$ will guide $f$ for feature refinement. Finally, we use $w^+$ and $f$ in StyleGAN to generate the reconstruction result.


We named our method CLCAE (i.e., StyleGAN inversion with \textbf{C}ontrastive \textbf{L}earning and \textbf{C}ross-\textbf{A}ttention \textbf{E}ncoder). We show that our CLCAE can achieve state-of-the-art performance in both reconstruction quality and editing capacity on benchmark datasets containing human portraits and cars. Fig.~\ref{fig:teaser} shows some results.  This indicates the robustness of our CLCAE. 
Our contributions are summarized as follows:
\begin{itemize}[noitemsep,nolistsep]
\item  We propose a novel contrastive learning approach to align the image space and foundation latent space $\mathcal{W}$ of StyleGAN. This alignment ensures that we can obtain proper latent code $w$ during GAN inversion.
\item We propose a cross-attention encoder to transform latent codes in $\mathcal{W}$ into $\mathcal{W^+}$ and $\mathcal{F}$. The representation of latent code in $\mathcal{W^+}$ and feature in $\mathcal{F}$ are improved to benefit reconstruction fidelity and editability.
\item Experiments indicate that our CLCAE achieves state-of-the-art fidelity and editability results both qualitatively and quantitatively.

\end{itemize}

\section{Related Work}


\subsection{GAN Inversion}
GAN inversion~\cite{zhu2016generative} is the task to find a latent code in a latent space of pretrained-GAN's domain for the real image. As mentioned in the GAN inversion survey~\cite{xia2021gan}, the inversion methods can be divided into three groups: optimization-based, encoder-based, and hybrid. The optimization-based methods~\cite{abdal2019image2stylegan,abdal2020image2stylegan++,creswell2018inverting,semantic2019bau,creswell2018inverting,gu2020image,yeh2017semantic,zhu2020improved} try to directly optimize the latent code or the parameters of GAN~\cite{roich2021pivotal} to minimize the distance between the reconstruction image. The encoder-based methods~\cite{richardson2020encoding,kang2021gan,Collins2020EditingIS,alaluf2021restyle,tov2021designing,guan2020collaborative,kim2021exploiting,hu2022style,luo2017learning,perarnau2016invertible}learn a mapper to transfer the image to the latent code. The hybrid methods~\cite{zhu2020domain,zhu2016generative} combine these two methods. 
{\flushleft \bf StyleGAN Inversion.}
Our work belongs to the StyleGAN inversion framework. Typically, there are three embedding spaces (i.e., $\mathcal{W}$~\cite{karras2019style}, $\mathcal{W^+}$~\cite{abdal2019image2stylegan}, and $\mathcal{F}$~\cite{kang2021gan}) and they are the trade-off design between the distortion and editability. The $\mathcal{W}$ is the foundation latent space of StyleGAN, several works~\cite{tov2021designing,zhu2020improved} have shown inverting the image into this space produces a high degree of editability but unsatisfied reconstruction quality. Differently, the $\mathcal{W^+}$ is developed from $\mathcal{W}$ to reduce distortions while suffering less editing flexibility.  On the other hand, the $\mathcal{F}$ space consists of specific features in SyleGAN, and these features are generated by the latent input code of foundation latent space $\mathcal{W}$ in the StyleGAN training domain. The $\mathcal{F}$ space contains the highest reconstruction ability, but it suffers the worst editability. Different from these designs that directly explore $\mathcal{W^+}$ and $\mathcal{F}$, we step back to explore $\mathcal{W}$ and use it to guide $\mathcal{W^+}$ and $\mathcal{F}$ to improve fidelity and editability.

\subsection{Latent Space Editing}
Exploring latent space's semantic directions improves editing flexibility. Typically, there are two groups of methods to find meaningful semantic directions for latent space based editing: supervised and unsupervised methods. The supervised methods~\cite{abdal2020styleflow,shen2020interpreting,denton2019detecting,goetschalckx2019ganalyze} need attribute classifiers or labeled data for specific attributes. InterfaceGAN~\cite{shen2020interpreting} use annotated images to train a binary Support Vector Machine~\cite{noble2006support} for each label and interprets the normal vectors of the obtained hyperplanes as manipulation direction. 
The unsupervised methods~\cite{voynov2020unsupervised,harkonen2020ganspace,shen2020closedform,yuksel2021latentclr} do not need the labels. GanSpace~\cite{harkonen2020ganspace} find directions use Principal Component Analysis (PCA). 
Moreover, some methods~\cite{hou2022feat, patashnik2021styleclip,wei2021hairclip,zhu2022one} use the CLIP loss~\cite{radford2021learning} to  achieve  amazing text guiding image manipulation. And some methods use the GAN-based pipeline to edit or inpaint the images~\cite{liu2019coherent,liu2020rethinking,liu2021deflocnet,liu2023human,liu2021pd}.In this paper, we follow the~\cite{tov2021designing} and use the InterfaceGAN and GanSpace to find the semantic direction and evaluate the manipulation performance.

\subsection{Contrastive Learning}
Contrastive learning~\cite{he2020momentum,grill2020bootstrap,ge2021revitalizing,pan2021videomoco,caron2021emerging,chongjian-snclr} has shown effective in self-supervised learning. When processing multi-modality data (i.e., text and images), CLIP~\cite{radford2021learning} provides a novel paradigm to align text and image features via contrastive learning pretraining. This cross-modality feature alignment motivates generation methods~\cite{kim2022diffusionclip,xia2021tedigan,patashnik2021styleclip,wei2021hairclip,zhu2022one} to edit images with text attributes. In this paper, we are inspired by the CLIP and align the foundation latent space $\mathcal{W}$ and the image space with contrastive learning. Then, we set the contrastive learning framework as a loss function to help us find the suitable latent code in $\mathcal{W}$ for the real image during GAN inversion.

\section{Method}

\begin{figure*}[t]
\begin{center}
\includegraphics[width=0.9\linewidth]{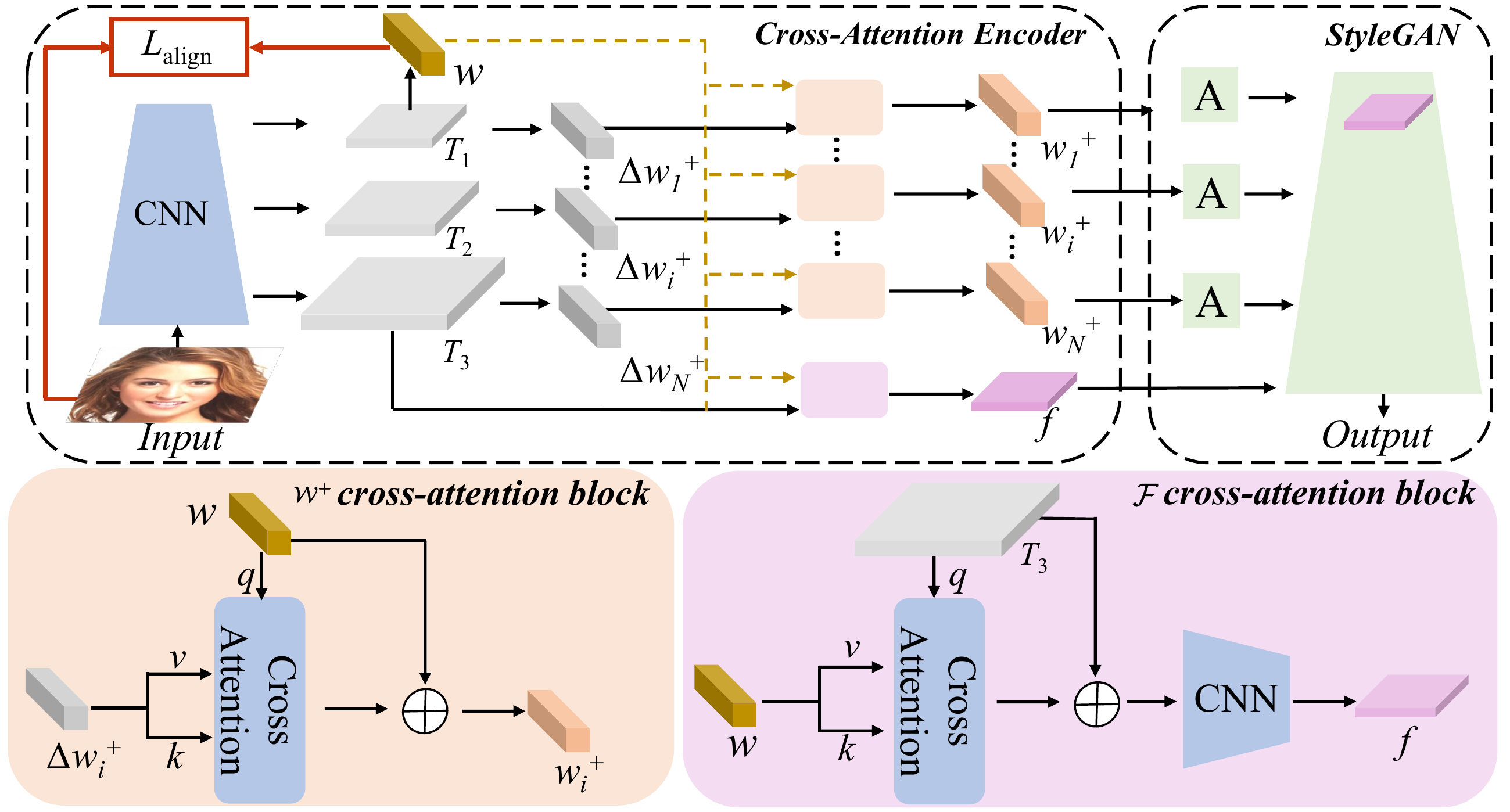}
\caption{The pipeline of our method. With the input image,  we first predict the latent code $w$  with feature $T_1$. The $w$ is constrained with the proposed $\mathcal{L}_{\text{align}}$. Then  two cross-attention blocks take the refined $w$ as a foundation to produce the latent code $w^+$ and feature $f$. Finally, we send the $w^+$ to StyleGAN via AdaIN~\cite{huang2017arbitrary} and replace the selected feature in StyleGAN with $f$ to generate the output image. }
\vspace{-0.1in}
\label{fig:pipeline}
\end{center}
\end{figure*}

Fig.~\ref{fig:pipeline} shows an overview of the proposed method. Our CNN encoder is 
from pSp~\cite{richardson2020encoding} that is the prevalent encoder in GAN inversion. Given an input image $I$, we obtain latent code $w$ in foundation latent space $\mathcal{W} \in \mathbb{R}^{512}$. This space is aligned to the image space via contrastive learning. Then we set the latent code $w$ as a query to obtain the latent code $w^+$ in $\mathcal{W^+} \in \mathbb{R}^{N \times 512} $ space via $\mathcal{W^+}$ cross-attention block. The size of $N$ is related to the size of the generated image (i.e., $N=18$ when the size of the generated image is $1024 \times 1024$). Meanwhile, we select the top feature in the encoder as $f$ in $\mathcal{F} \in \mathbb{R}^{H \times W \times C}$ space and use $w$ to refine $f$ with $\mathcal{F}$ cross-attention block. Finally, we send $w^+$ and $f$ to the pretrained StyleGAN pipeline to produce the reconstruction results. 


\begin{figure}[t]
\begin{center}
\includegraphics[width=1\linewidth]{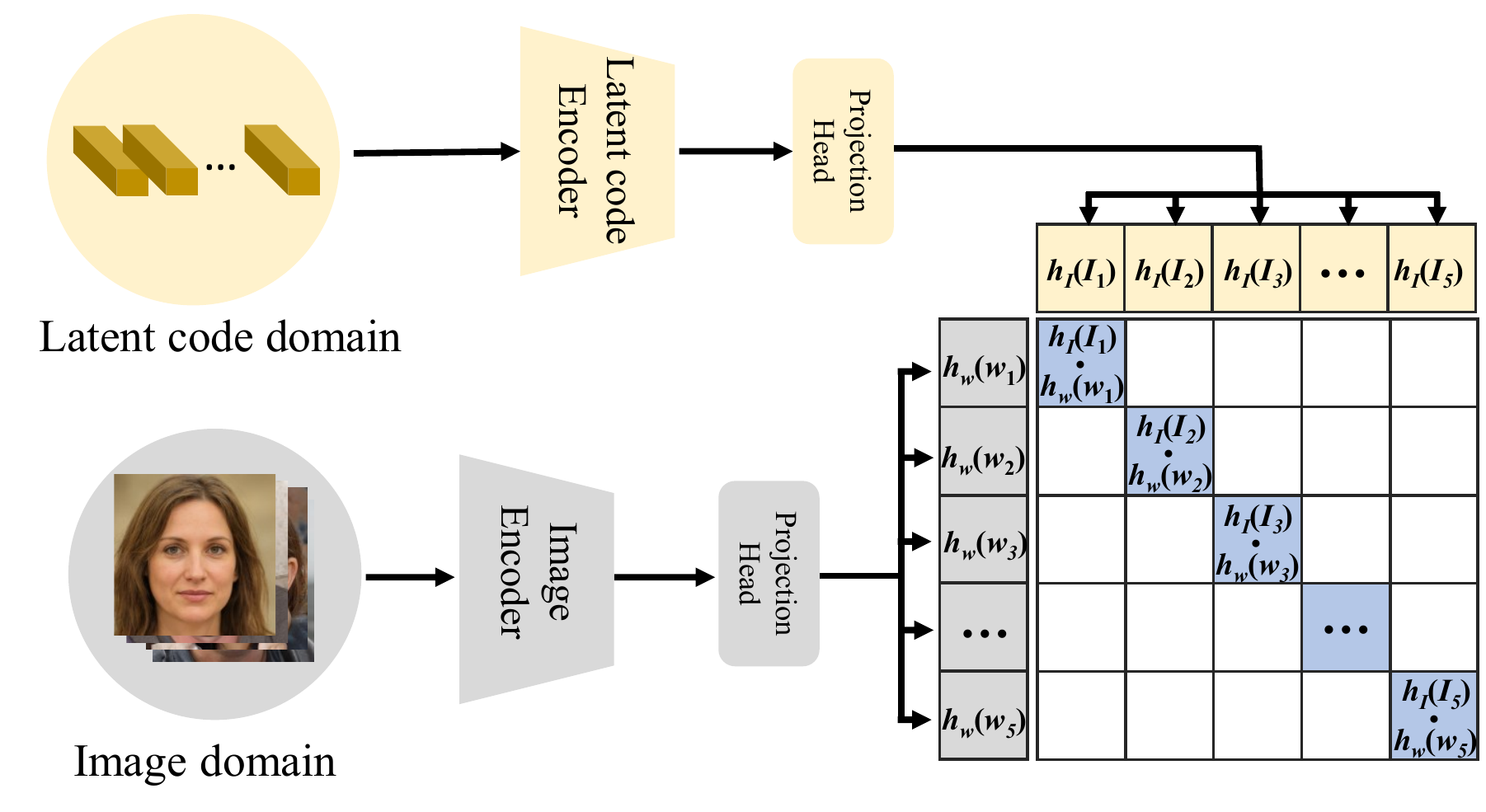}

\caption{The process of contrastive learning pre-training. The encoders and projection heads extract the embedding of the image and latent code. Then we make the paired embeddings similar to align the image and latent code distribution. After alignment, we fix the parameters in the contrastive learning module to enable the latent code to fit the image during inversion.}
\vspace{-0.2in}
\label{fig:cl}
\end{center}
\end{figure}

\subsection{Aligning Images and Latent Codes}
We use contrastive learning from CLIP to align image $I$ and its latent code $w$. After pre-training, we fix this module and use it as a loss function to measure the image and latent code similarity. This loss is set to train the CNN encoder in Fig.~\ref{fig:pipeline} as to align one image $I$ and its latent code $w$.

The contrastive learning module is shown in Fig.~\ref{fig:cl}. We 
synthesize 100K ($I$) and latent code($w$) pairs with pretrained StyleGAN. The $I$ and $w$ are fed into the module where there are feature extractors (i.e., CNN for $I$ and transformer for $w$) and projection heads. Specifically, our minibatch contains $S$ image and latent code pairs ($I\in \mathbb{R}^{256 \times 256 \times 3}$, $w\in \mathbb{R}^{512}$). We denote their embeddings after projection heads (i.e., hidden state) as $h_I(I) \in \mathbb{R}^{512}$ and $h_w(w) \in \mathbb{R}^{512}$, respectively. For the $i$-th pair from one minibatch (i.e., $i\in[1,2,...,S]$), its embeddings are $h_I(I_i)$ and $h_w(w_i)$. The contrastive loss~\cite{zhang2020contrastive,oord2018representation} can be written as
\begin{equation}
\mathcal{L}_i^{(I \rightarrow w)}=-\log \frac{\exp \Big[\left\langle h_I(I_i), h_w(w_i)\right\rangle / t \Big]}{\sum_{k=1}^S \exp \Big[\left\langle h_I(I_i), h_w(w_k)\right\rangle / t \Big]},
\end{equation}
\begin{equation}
\mathcal{L}_i^{(w \rightarrow I)}=-\log \frac{\exp \Big[\left\langle h_w(w_i), h_I(I_i)\right\rangle / t \Big]}{\sum_{k=1}^S \exp \Big[\left\langle h_w(w_i),  h_I(I_k)\right\rangle / t \Big]},
\end{equation} 
where $\left\langle \cdot \right\rangle$ denotes the cosine similarity, and $ t \in \mathbb{R}^{+}$ is a learnable temperature parameter. The alignment loss in the contrastive learning module can be written as
\begin{equation}\label{eq:align}
\mathcal{L}_{\text{align}} =\frac{1}{S} \sum_{i=1}^S\left(\lambda \mathcal{L}_i^{(I \rightarrow w)}+(1-\lambda) \mathcal{L}_i^{(w \rightarrow I)}\right),
\end{equation}
where $\lambda = 0.5$. We use the CNN in pSp~\cite{richardson2020encoding}  as the image encoder, and StyleTransformer \cite{hu2022style} as the latent code encoder. Then in the GAN inversion process, we fix the parameters in the contrastive learning module and compute $\mathcal{L}_{align}$ to enable the latent code to fit the image. Aligning images to their latent codes directly via supervision $\mathcal{L}_{align}$ enforces our foundation latent space $\mathcal{W^+}$ close to the image space to avoid reconstruction distortions.

\subsection{Cross-Attention Encoder}

Once we have pre-trained the contrastive learning module, we make it frozen to provide the image and latent code matching loss. This loss function is utilized for training the CNN encoder in our CLCAE framework shown in Fig.~\ref{fig:pipeline}. Our CNN encoder is a pyramid structure for hierarchical feature generations (i.e., $T_1, T_2, T_3$). We use $T_1$ to generate latent code $w$ via a map2style block. Both the CNN encoder and the map2style block are from pSp~\cite{richardson2020encoding}. After obtaining $w$, we can use $I$ and $w$ to produce an alignment loss via Eq.~\ref{eq:align}. This loss will further update the CNN encoder for image and latent code alignment. Also, we use $w$ to discover $w^+$ and $f$ with the cross-attention blocks.



\subsubsection{$\mathcal{W^+}$ Cross-Attention Block}
As shown in Fig.~\ref{fig:pipeline}, we set the output of  $\mathcal{W^+}$ cross-attention block as the residual of $w$ to predict   $w^+$. Specifically, we can get the coarse residual $\Delta w^+\in \mathbb{R}^{N  \times 512}$ with the CNN's features and map2style blocks first. Then we send each vector $\Delta w^+_i\in \mathbb{R}^{512}$ in $\Delta w^+$ and $w\in \mathbb{R}^{512}$ to the $\mathcal{W^+}$ cross-attention block to predict the better $\Delta w^+_i$, where $i =  1,...,N $. In the cross-attention block, we set the $w$ as query($Q$) and $\Delta w^+_i$ as value($V$) and key($K$) to calculate the attention map. This attention map can extract the potential relation between the $w$ and $\Delta w^+_i$, and it can make the $w^+$ close to the $w$. Specifically, the $Q$, $K$, and $V$ are all projected from $\Delta w^+_i$ and $w$ with learnable projection heads, and we add the output of cross-attention with $w$ to get final latent code $w^+_i$ in $\mathcal{W^+}$, the whole process can be written as
\begin{equation}\label{eq:wplusatt}
\begin{aligned}
& Q=w W_{Q}^{w^+}, K = \Delta w^+_i W_{K}^{w^+},  V = \Delta w^+_i   W_{V}^{w^+}, \\
& \operatorname{Attention}(Q, K, V) = \operatorname{Softmax}\left(\frac{Q K^T}{\sqrt{d}}\right) V, \\
& w^+_i = w + \operatorname{Attention}(Q, K, V),
\end{aligned}
\end{equation}
where $W_{Q}^{w^+}$,  $W_{K}^{w^+}$,  $W_{V}^{w^+} \in {R}^{512 \times 512}$ and the feature dimension $d$ is 512. We use the multi-head mechanism~\cite{vaswani2017attention} in our cross-attention. The cross-attention can make the $w^+$ close to the $w$ to preserve the great editability. Meanwhile, the reconstruction performance can still be preserved, since we get the refined $w$ via the $\mathcal{L}_{\text{align}}$.

 \subsubsection{$\mathcal{F}$ Cross-Attention Block}  
The rich and correct spatial information can improve the representation ability of $f$ as mentioned in ~\cite{parmar2022spatially}. We use the $T_3\in \mathbb{R}^{64\times64\times512}$ as our basic feature to predict $f$ as shown in Fig.~\ref{fig:pipeline}, since the $T_3$ has the richest spatial information in the pyramid CNN. Then we calculate cross attention between the $w$ and  $T_3$  and output a residual to refine the $T_3$. In contrast to the $ W^+ $ cross-attention block, we set the $w$ as value($V$)  and key($K$)  and $T_3$ as query($Q$), this is because we want to explore the spatial information of $w$ to support the $T_3$. Finally, we  use a CNN to reduce the spatial size of the cross-attention block's output to get the final prediction $f$, the shape of $f$ matches the feature of the selected  convolution layer in $\mathcal{F}$ space. We choose the 5th convolution layer following the FS~\cite{xuyao2022}. The whole process can be written as:
\begin{equation}\label{eq:fatt}
\begin{aligned}
& Q=T_3 W_{Q}^{f}, K =  w W_{K}^{f},  V =   w   W_{V}^{f}, \\
& \operatorname{Attention}(Q, K, V) = \operatorname{Softmax}\left(\frac{Q K^T}{\sqrt{d}}\right) V,\\
& f = \operatorname{CNN}\Big[\operatorname{Attention}(Q, K, V) + T_3\Big],
\end{aligned}
\end{equation}
where $W_{Q}^{f}$,  $W_{K}^{f}$,  $W_{V}^{f} \in {R}^{512 \times 512}$ and the feature dimension $d$ is 512. Finally, we send the $ w^+ $ to the pretrained StyleGAN ($G$) via AdaIN~\cite{huang2017arbitrary} and replace the selected feature in $G$ with $f$ to get the final reconstruction result $G(w^+,f)$.  

{\flushleft \bf Image Editing.} 
During the editing process, we need to get the modified ${\hat{w}^+}$ and $\hat{f}$. For the ${\hat{w}^+}$, we obtain it with the classic latent space editing methods~\cite{harkonen2020ganspace,shen2020interpreting}. For the $\hat{f}$, we follow the FS~\cite{xuyao2022} to generate the reconstruction result $G(w^+)$ and the edited image $G(\hat{w^+})$ respectively first. Then we extract the feature of the 5th convolution layer of $G(\hat{w^+})$ and  $G( {w^+})$ respectively. Finally, we calculate the difference between these two features and add it to the  ${f}$ to predict the $\hat{f}$. The whole process to get the $\hat{f}$ is:
\begin{equation}
\hat{f}=f+G^5(\hat{ w})- G^5(w),
\label{fedit}
\end{equation} where the $ {G}^5(\tilde{ {w}}) $ and ${G}^5( {w})$ is the feature of 5-th convolution layer. With the modified ${\hat{w}^+}$ and $\hat{f}$, we can get the editing results $G({\hat{w}^+}, \hat{f})$.

 \subsection{Loss Functions}
To train our encoder, we use the common ID and reconstruction losses  to optimize the three reconstruction results $I_{rec}^1 = G(w)$, $I_{rec}^{2}=G(w^+)$ and $I_{rec}^{3}=G(w^+, f)$ simultaneously. Meanwhile, we use the feature regularization to make the $f$ close to the original feature in $G$ similar to the  FS~\cite{xuyao2022}.  

{\flushleft \bf Reconstruction losses.} We utilize the pixel-wise $\mathcal{L}_2$ loss and $\mathcal{L}_{\text {LPIPS }}$~\cite{zhang2018perceptual} to measure the pixel-level and perceptual-level similarity between the input image and reconstruction image as
\begin{equation}
\mathcal{L}_{rec}=\sum_{i=1}^3 \left(\lambda_{\text {LPIPS }} \mathcal{L}_{\text {LPIPS }}\left(I, I_{rec}^i\right)+ \lambda_{2}\mathcal{L}_2\left(I, I_{rec}^i\right)\right),
\end{equation}where the $\mathcal{L}_{\text {LPIPS }}$ and  $\mathcal{L}_{2}$ are are weights balancing each loss. We set the $\mathcal{L}_{\text {LPIPS }} = 0.2$ and  $\mathcal{L}_2=1$ during training.
{\flushleft \bf ID loss.} 
We follow the e4e~\cite{tov2021designing} to use the identity loss to preserve the identity of the reconstructed image as
\begin{equation}
\mathcal{L}_{i d} =\sum_{i=1}^3\left(1-\left\langle\text{R}({I}), \text{R}(I_{rec}^i)\right\rangle\right).
\end{equation} For the human portrait dataset, the $\text{R}$ is a pretrained ArcFace facial recognition network~\cite{huang2020curricularface}. For the cars dataset, the $\text{R}$ is a ResNet-50~\cite{he2015deep} network trained with MOCOv2~\cite{chen2020improved}.

 {\flushleft \bf Feature regularization.} 
To edit the $f$ with Eq.~\ref{fedit}, we need to ensure $f$ is similar to the original feature of $G$. So we adopt a regularization for the $f$ as
\begin{equation}
\mathcal{L}_{f_  \text {reg }}=\left\|f- {G}^5(w^+)\right\|_2^2.
\end{equation}

 {\flushleft \bf Total losses.}  In addition to the above losses, we add the $\mathcal{L}_{\text{align}}$ to help us find the proper $w$. In summary, the total loss function is defined as:
 \begin{equation}
 \begin{gathered}
\mathcal{L}_{total} =\lambda_{\text {rec }}\mathcal{L}_{\text {rec }} +\lambda_{\text {ID}} \mathcal{L}_{\text {ID}} + \lambda_{f_\text {reg }}\mathcal{L}_{f_  \text {reg }} +
\lambda_{\text {align}}\mathcal{L}_{\text{align}},
\end{gathered}
\end{equation}
where $\lambda_{\text {rec }}$, $\lambda_{\text {ID}}$, $\lambda_{f_\text {reg }}$ and $\lambda_{\text {align}}$ are the weights that adjust the contribution of each loss term. And we set the $\lambda_{\text {rec }} =1 $, $\lambda_{\text {ID}} = 0.1$, $\lambda_{f_\text {reg }=0.01}$ and $\lambda_{\text {align}}=1$ respectively by default.

\begin{figure*}[t]
\begin{center}
\includegraphics[width=0.9\linewidth]{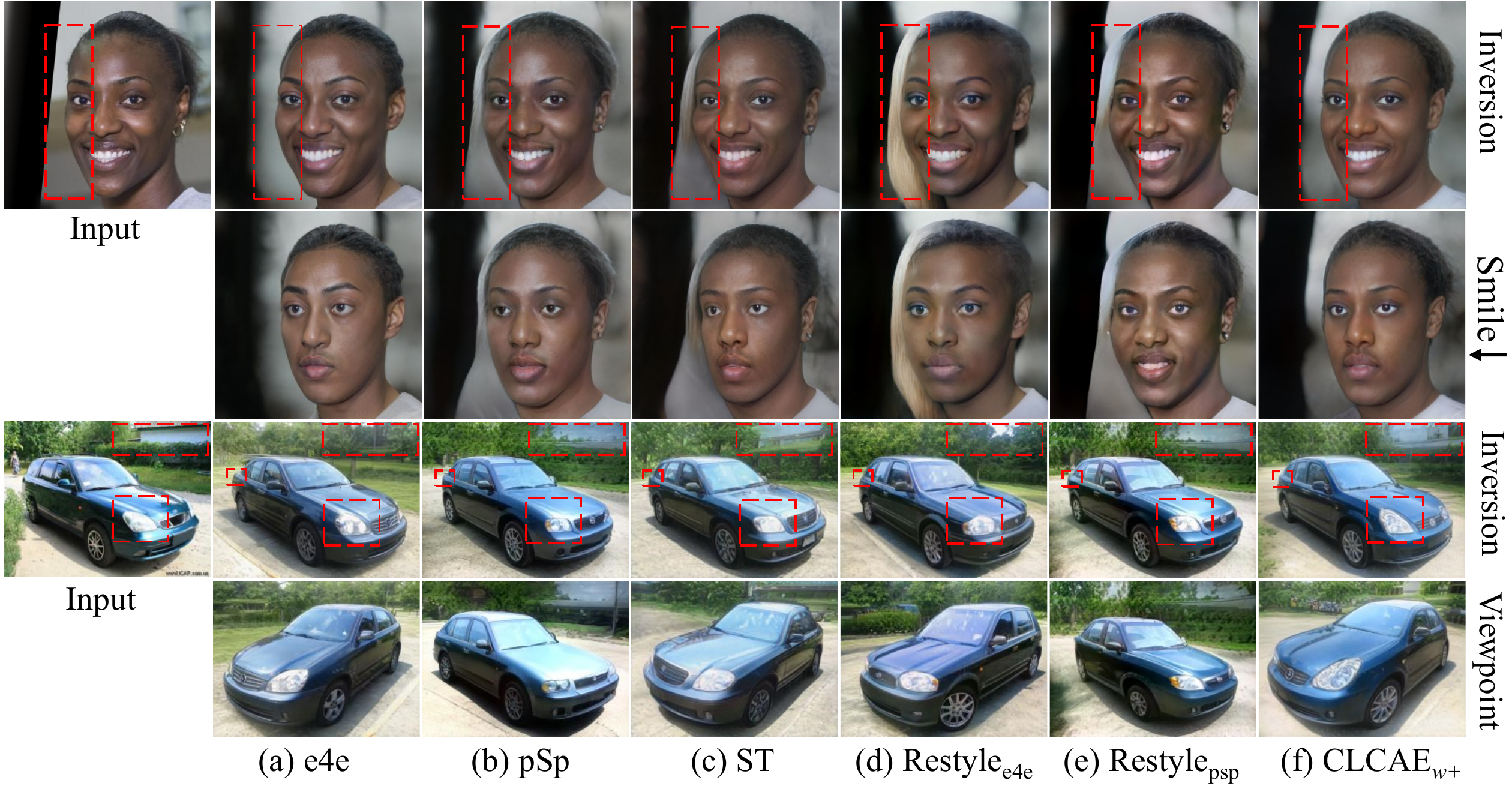}
\caption{   Visual comparison of inversion and editing between our method and the baseline methods (e4e~\cite{tov2021designing}, pSp~\cite{richardson2020encoding}, ST~\cite{hu2022style}, restyle$_{\text{e4e}}$~\cite{alaluf2021restyle} and restyle$_{\text{pSp}}$~\cite{alaluf2021restyle}) in the $\mathcal{W^+}$ group. We produce CLCAE$_{w^+} = G(w^+)$ to compare with them.  Our method is more effective in producing manipulation attribute  relevant and visually realistic results. $\downarrow$ means a reduction of the manipulation attribute.}
\vspace{-0.1in}
\label{fig:wplusresults}
\end{center}

\end{figure*}

\section{Experiments}
In this section, we first illustrate our implementation details. Then we compare our method with existing methods qualitatively and quantitatively. Finally, an ablation study validates the effectiveness of our contributions.  More results are provided in the supplementary files. We will release our implementations to the public.

\subsection{Implementation Details}

During the contrastive learning process, we follow the CLIP~\cite{radford2021learning} and use the Adam optimizer~\cite{kingma2014adam} to train the image and latent code encoders. We synthesize the image-latent code pair dataset with the pre-trained StyleGAN2 in cars and human portrait domains. We set the batch size to 256 for training. During the StyleGAN inversion process, we train and evaluate our method on cars and human portrait datasets. For the human portrait, we use the FFHQ~\cite{karras2019style} dataset for training and the CelebA-HQ test set~\cite{liu2015deep} for evaluation. For cars, we use the Stanford Cars~\cite{KrauseStarkDengFei-Fei_3DRR2013} dataset for training and testing. We set the resolution of the input image as $256 \times 256$. We follow the pSp~\cite{richardson2020encoding} and use the Ranger optimizer to train our encoder for GAN inversion, the Ranger optimizer is a combination of Rectified Adam~\cite{liu2019variance} with the Lookahead technique~\cite{zhang2019lookahead}. We set the batch size to 32 during training. We use  8 Nvidia Telsa V100 GPUs to train our model.

\begin{figure*}[t]
\begin{center}
\includegraphics[width=0.9\linewidth]{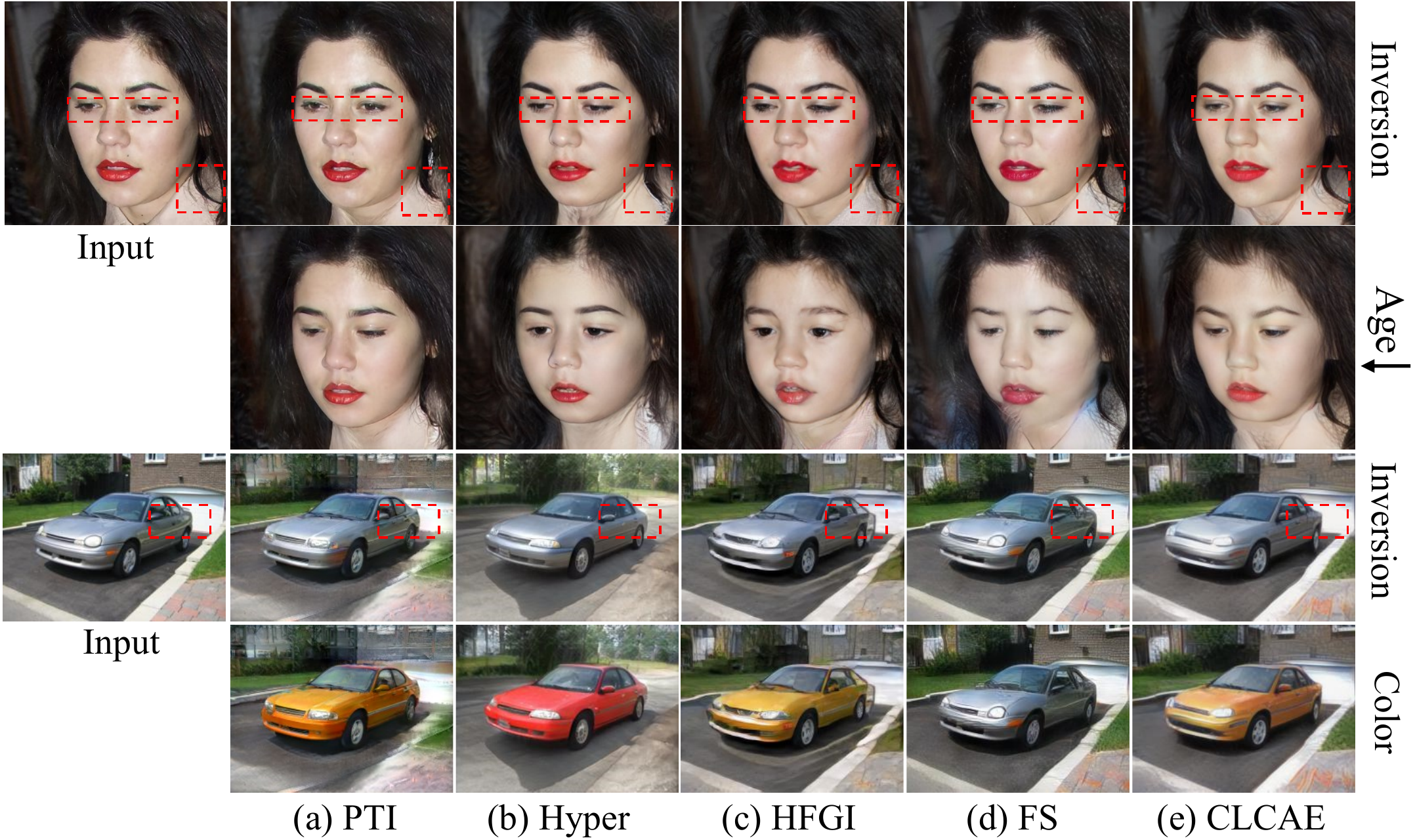}
\caption{Visual comparison of inversion and editing between our method and the baseline methods (PTI~\cite{roich2021pivotal}, Hyper~\cite{alaluf2022hyperstyle}, HFGI~\cite{wang2022high}, and FS~\cite{xuyao2022}) in the $\mathcal{F}$ group. We produce CLCAE = $G(w^+, f)$ to compare with them. Our method not only generates high-fidelity reconstruction results but also retains the flexible manipulation ability.  $\downarrow$ means a reduction of the manipulation attribute. }
\vspace{-0.2in}
\label{fig:fresults}
\end{center}
\end{figure*}

 \begin{figure*}[t]
\begin{center}
\includegraphics[width=0.85\linewidth]{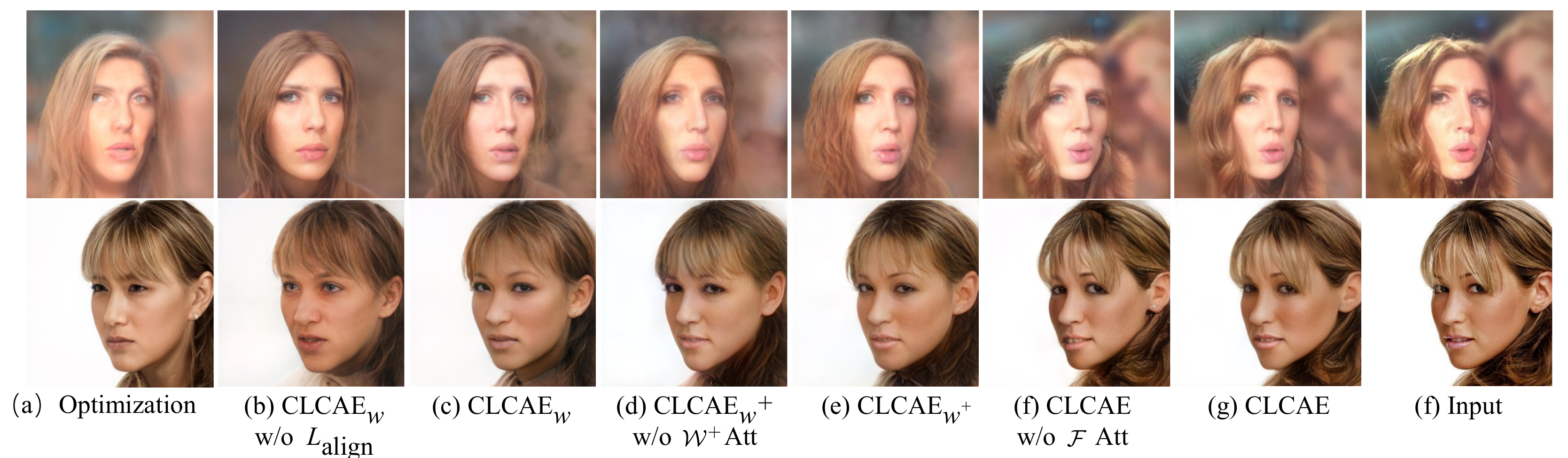}
\caption{Visual results of ablation study. The (a) is an Optimization~\cite{karras2020analyzing} method which inverts the image to the $\mathcal{W}$ space. The (b) and (c) are the results generated by $w$ with and without  $\mathcal{L}_{\text{align}}$ respectively. By comparing (a), (b), and (c), we can see that $\mathcal{L}_{\text{align}}$ can  help our method produce better latent code $w$ than optimization-based methods. (c) and (d) are the results generated by $w^+$ with and without $\mathcal{W^+}$ cross-attention block  respectively. The (e) and (f) are the results generated by both $w^+$ and $f$ with and without $\mathcal{F}$ cross-attention block, respectively. The performance gap between every two results can prove the effectiveness of $w^+$ and $f$ cross-attention blocks.}
\vspace{-0.1in}
\label{fig:ablation}
\end{center}
\end{figure*}

\subsection{Qualitative Evaluation}
Our CLCAE improves the representation ability of the latent code in $\mathcal{W^+}$ and feature in $\mathcal{F}$ spaces. We evaluate qualitatively how our latent codes $w^+$ and $f$ improve the output result. To clearly compare these two latent codes, we split the evaluation methods into two groups. The first group consists of methods only using latent code $w+$, we denote this group as `group $\mathcal{W^+}$'. The second group consists of methods using both $w+$ and $f$, we denote this group as `group $\mathcal{F}$'. When comparing to the group $\mathcal{W^+}$, we use our results CLCAE$_{w^+}$ computed via $G(w^+)$ for fair comparisons. When comparing to the group $\mathcal{F}$, we use our results computed via $G(w^+, f)$. During image editing, we use InterfaceGAN\cite{shen2020interpreting} and GanSpace~\cite{harkonen2020ganspace} to find the semantic direction and manipulate the face and car images, respectively.


{\flushleft \bf $\mathcal{W^+}$ space.}
 Fig.~\ref{fig:wplusresults} shows the visual results where our CLCAE$_{w^+}$ is compared to e4e~\cite{tov2021designing}, pSp~\cite{richardson2020encoding}, restyle$_{\text{pSp}}$~\cite{alaluf2021restyle}, restyle$_{\text{e4e}}$~\cite{alaluf2021restyle} and  StyleTransformer (ST)~\cite{hu2022style}. 
Both our CLCAE$_{w^+}$ and e4e show better inversion performance in the human portrait. This phenomenon is caused by the overfitting of those methods in (b)$ \sim $ (e), since the $\mathcal{W^+}$ space pays more attention to the quality of the reconstruction. The CLCAE$_{w^+}$ and e4e can produce $w^+$ close to the $w$, which improves the robustness of these two methods. Moreover, our CLCAE$_{w^+}$ is more capable of avoiding distortions while maintaining editability than other methods, including e4e (see the second row). This is because our $w^+$ is based on the solid  $w$  that does not damage the reconstruction performance of $w^+$. For the domain of cars, we observe that pSp and restyle$_{\text{pSp}}$   are limited to represent editing ability (see the (b) and (e) of the viewpoint row). On the other hand, e4e and ST are able to edit images, but their reconstruction performance are unsatisfying. In contrast to these methods, our CLCAE$_{w^+}$ maintains high fidelity and flexible editability at the same time.

{\flushleft \bf $\mathcal{F}$ space.} Fig.~\ref{fig:fresults} shows our comparisons to PTI~\cite{roich2021pivotal}, Hyper~\cite{alaluf2022hyperstyle}, HFGI~\cite{wang2022high}, and FS~\cite{xuyao2022}  in the $\mathcal{F}$ space. The results of PTI, Hyper, HFGI, and FS contain noticeable distortion in the face  (e.g., the eyes in the red box regions in (a)$ \sim $ (d)) and the car (e.g., the background in (a)$ \sim $ (c) and the red box regions in car images). Although FS~\cite{xuyao2022} reconstructs the background of the car image well, it loses editing flexibility (e.g., see (d) of  4 rows). This is because the  FS method relies too much on $\mathcal{F}$ space, which limits the editability. In contrast, our results are in high fidelity as well as a wide range of editability with powerful $f$ and $w^+$.



\begin{table*}[t]
    \Huge
	\renewcommand\arraystretch{0.5}
	\centering  
	\caption{ Quantitative comparisons of state-of-the-art methods on the CelebA-HQ dataset.  We conduct a user study to measure the editing performance. The number denotes the preference rate of our method against the competing methods. Chance is 50\%.  ${\downarrow}$ indicates lower is better while ${\uparrow}$ indicates higher is better. }  
\resizebox{0.9\linewidth}{!}{
\renewcommand{\arraystretch}{1}
\begin{tabular}{cc|cccccc|ccccc}
\hline
\multicolumn{2}{c|}{Group}                                 & \multicolumn{6}{c|}{$\mathcal{W^+}$}                                          & \multicolumn{5}{c}{$\mathcal{F}$}                                     \\ \hline
\multicolumn{2}{c|}{Method}                                & e4e~\cite{tov2021designing}   & pSp~\cite{richardson2020encoding}   & TS~\cite{hu2022style}    & restyle$_{\text{e4e}}$~\cite{alaluf2021restyle} & restyle$_{\text{pSp}}$~\cite{alaluf2021restyle}       & CLCAE$_{w^+}$          & PTI~\cite{roich2021pivotal}           & Hyper~\cite{alaluf2022hyperstyle}   & HFGI~\cite{wang2022high}  & FS~\cite{xuyao2022}    & CLCAE       \\ \hline
\multicolumn{1}{c|}{}                         & PSNR$\uparrow$         & 19.08 & 20.39 & 20.50 & 19.45   & 21.20         & \textbf{21.23} & 23.49         & 22.09    & 22.13 & 24.08 & \textbf{24.50} \\ \cline{2-2}
\multicolumn{1}{c|}{}                         & SSIM$\uparrow$        & 0.53  & 0.56  & 0.57  & 0.54    & 0.57          & \textbf{0.59}  & 0.65          & 0.61     & 0.62  & 0.67  & \textbf{0.68}  \\ \cline{2-2}
\multicolumn{1}{c|}{Inversion}                & LPIPS$\downarrow$      & 0.20  & 0.16  & 0.16  & 0.19    & \textbf{0.13} & 0.15           & 0.09          & 0.10     & 0.12  & 0.07  & \textbf{0.06}  \\ \cline{2-2}
\multicolumn{1}{c|}{}                         & ID$\uparrow$         & 0.50  & 0.56  & 0.59  & 0.50    & 0.65          & \textbf{0.65}  & \textbf{0.83} & 0.74     & 0.68  & 0.75  & 0.79           \\ \cline{2-2}
\multicolumn{1}{c|}{}                         & Time$\downarrow$        & 0.029s & 0.028s & 0.026s & 1.154s   & 1.150s         & 0.071s          & 355.323s       & 1.161s    & 0.036s & 0.581s & 0.080s          \\ \hline
\multicolumn{1}{c|}{\multirow{2}{*}{Editing}} & ID$\uparrow$ (Smile)         & 0.44  & 0.52  & 0.53   &0.47   &\textbf{0.64}           & 0.62               & 0.57         &0.62    &  0.54     &0.66       &\textbf{0.67}                \\ \cline{2-2}
\multicolumn{1}{c|}{}                         & User Study$\downarrow$ & 70\% & 60\% & 62\% & 84\%   & 73\%         & -              & 74\%         & 72\%    & 60\% & 96\% & -              \\ \hline
\end{tabular}

}
\label{tab:quantitative}
\end{table*}

\subsection{Quantitative Evaluation}
     {\flushleft \bf Inversion.} We perform a quantitative comparison in the CelebA-HQ dataset to evaluate the inversion performance. We apply the commonly-used metric: PSNR, SSIM, LPIPS~\cite{zhang2018perceptual} and ID~\cite{huang2020curricularface}. Table~\ref{tab:quantitative} shows these evaluation results. The PTI in $\mathcal{F}$ group and Restyle$_{pSp}$ in $\mathcal{W^+}$ group have better performance than our method in ID and LPIPS metric, respectively.  But these two method  takes a lot of time for the optimization operation or the iterative process. With the simple and effective cross-attention encoder and the proper foundation latent code, our method can achieve good performance in less time. 
 {\flushleft \bf Editing.}
There is hardly a straight quantitative measurement to evaluate editing performance. We use the InterFaceGAN~\cite{shen2020interpreting} to find the manipulation direction and edit the image, then we calculate the ID distance~\cite{huang2020curricularface} between the original image and the manipulation one. For a fair comparison, during the ID distance evaluation,  we use the "smile" manipulation direction and adopt the same editing degree for CLCAE and other baselines. Besides using the object metric to evaluate the editing ability, we conduct a user study on the manipulated results from compared methods. We randomly collected 45 images of faces and cars for 9 groups of comparison methods, each group has 5 images, and these images are edited with our method and a baseline method, respectively. 20 participants need to select the one edited image with higher fidelity and proper manipulation. The user study is shown in Table~\ref{tab:quantitative}. The results indicate that most participants support our approach.

\subsection{Ablation Study}
 
{\flushleft \bf Effect of contrastive learning.} We compare the  optimization method~\cite{karras2020analyzing}   to evaluate whether our method can predict the solid latent code in foundation $\mathcal{W}$ space. The optimization method (a) can invert the image to the $\mathcal{W}$ with a fitting process. The visual comparisons are shown in Fig.~\ref{fig:ablation}, CLCAE$_{w}$ in (c) is the reconstruction results generated with our latent code $w$. Our method outperforms the optimization method in the ability of reconstruction and identity preservation. This is because the proposed $L_{\text{align}}$ can directly calculate the distance between the latent code $w$ and the image, while the optimization method only measures the difference in the image domain. Meanwhile, we present the results generated by $w$ without $L_{\text{align}}$ in (b) to prove our contrastive learning validity further. The associated numerical results are shown in Table \ref{tab:ablation}.
{\flushleft \bf Effect of the $\mathcal{W^+}$ Cross-Attention. }
To validate the effectiveness of $\mathcal{W^+}$ cross-attention block, we remove it and use the coarse residual as $w^+$ directly to do a comparison experiment. As shown in Fig.~\ref{fig:ablation}, the experiment results in (d) have distortion (see the eyes regions of the first row and the hair regions of the second row). And the cross-attention block in (e) can improve performance. This is because the cross-attention block utilizes the solid  latent code $w$ to support our method to predict better  $w^+$.  The numerical analysis results are shown in Table \ref{tab:ablation}.

 \begin{table}[t] 
    \Huge
	\centering  
	\caption{ Quantitative ablation study on the CelebA-HQ dataset. ${\downarrow}$ indicates lower is better while ${\uparrow}$ indicates higher is better.}
\resizebox{1\linewidth}{!}{
\renewcommand{\arraystretch}{1.4}

\begin{tabular}{|c|c|c|c|c|c|c|c|}  
\hline
Method & \begin{tabular}[c]{@{}c@{}}Optimization \\ ~\cite{karras2020analyzing} \end{tabular} & \begin{tabular}[c]{@{}c@{}}CLCAE$_w$\\ w/o $\mathcal{L}_{\text{align}}$ \end{tabular} & CLACE$_w$       & \begin{tabular}[c]{@{}c@{}}CLCAE$_{w^+}$ \\ w/o  $\mathcal{W^+}$ Att\end{tabular} & CLCAE$_{w^+}$ & \begin{tabular}[c]{@{}c@{}}CLCAE  \\ w/o $\mathcal{F}$ Att\end{tabular} & CLCAE                                \\ \hline
PSNR$\uparrow$   & 16.95        & 18.15 & \textbf{19.36}      & 20.61       & \textbf{21.23} & 23.93   & \textbf{24.50}          \\ 
SSIM$\uparrow$   & 0.53         & 0.52  & \textbf{0.54}    & 0.57       & \textbf{0.59}  & 0.66      & \textbf{0.679}             \\ 
LPIPS$\downarrow$  & 0.23         & 0.26  & \textbf{0.22}      & 0.20   & \textbf{0.15}  & 0.10    & \textbf{0.06}               \\ 
ID$\uparrow$      & 0.19         & 0.26  & \textbf{0.50}      & 0.56   & \textbf{0.65}  & 0.70     & \textbf{0.79}                \\ 
Time$\downarrow$   & 193.50s       &0.022s       & 0.022s     &0.028s      &0.071s      &0.074s   &0.080s                                                                                   \\ \hline
\end{tabular}
}
\label{tab:ablation}
\end{table}

{\flushleft \bf Effect of the $\mathcal{F}$ Cross-Attention. }
We analyze the effect of $\mathcal{F}$ cross-attention block by comparing the results produced with it and without it. We can see the visual comparison in Fig.~\ref{fig:ablation}. The results in (f) show that our method has artifacts in the hair and eye regions of the face  without the $\mathcal{F}$ cross-attention block. And our method with $\mathcal{F}$ cross-attention block demonstrates better detail (see the hair and eyes in (g)). This phenomenon can prove that the $\mathcal{F}$ cross-attention block can extract the valid information in $w$ and refine the $f$, which also tells us the importance of a good foundation. The numerical evaluation in Table \ref{tab:ablation} also indicates that $\mathcal{F}$ cross-attention block improves the quality of reconstructed content.

\section{Conclusion and Future Work}
we propose a novel GAN inversion method CLCAE that revisits the StyleGAN inversion and editing from the foundation space $\mathcal{W}$ viewpoint. CLCAE adopts a contrastive learning pre-training to align the image space and latent code space first. And we formulate the pre-training process as a loss function $\mathcal{L}_{\text{align}}$ to optimize   latent code $w$ in $\mathcal{W}$ space during inversion. Finally, CLCAE sets the $w$ as the foundation to obtain the proper $w^+$ and $f$ with proposed cross-attention blocks. Experiments on human portrait and car datasets prove that our method can simultaneously produce powerful $w$, $w^+$, and $f$. In the future, we will try to expand this contrastive pre-training process to other domains (e.g., Imagenet dataset~\cite{deng2009imagenet}) and 
do some basic downstream tasks such as classification and segmentation. This attempt could bring a new perspective to contrastive learning.

{\small
\bibliographystyle{ieee_fullname}
\bibliography{egbib}

\begin{thebibliography}{10}\itemsep=-1pt

\bibitem{abdal2019image2stylegan}
Rameen Abdal, Yipeng Qin, and Peter Wonka.
\newblock Image2stylegan: How to embed images into the stylegan latent space?
\newblock In {\em Proceedings of the IEEE international conference on computer
  vision}, 2019.

\bibitem{abdal2020image2stylegan++}
Rameen Abdal, Yipeng Qin, and Peter Wonka.
\newblock Image2stylegan++: How to edit the embedded images?
\newblock In {\em Proceedings of the IEEE/CVF Conference on Computer Vision and
  Pattern Recognition}, 2020.

\bibitem{abdal2020styleflow}
Rameen Abdal, Peihao Zhu, Niloy Mitra, and Peter Wonka.
\newblock Styleflow: Attribute-conditioned exploration of stylegan-generated
  images using conditional continuous normalizing flows, 2020.

\bibitem{alaluf2021matter}
Yuval Alaluf, Or Patashnik, and Daniel Cohen-Or.
\newblock Only a matter of style: Age transformation using a style-based
  regression model.
\newblock {\em ACM Trans. Graph.}, 2021.

\bibitem{alaluf2021restyle}
Yuval Alaluf, Or Patashnik, and Daniel Cohen-Or.
\newblock Restyle: A residual-based stylegan encoder via iterative refinement.
\newblock In {\em Proceedings of the IEEE/CVF International Conference on
  Computer Vision}, October 2021.

\bibitem{alaluf2022hyperstyle}
Yuval Alaluf, Omer Tov, Ron Mokady, Rinon Gal, and Amit Bermano.
\newblock Hyperstyle: Stylegan inversion with hypernetworks for real image
  editing.
\newblock In {\em Proceedings of the IEEE/CVF Conference on Computer Vision and
  Pattern Recognition}, 2022.

\bibitem{semantic2019bau}
David Bau, Hendrik Strobelt, William Peebles, Jonas Wulff, Bolei Zhou, Jun-Yan
  Zhu, and Antonio Torralba.
\newblock Semantic photo manipulation with a generative image prior.
\newblock {\em arXiv preprint arXiv:2005.07727}, 2020.

\bibitem{caron2021emerging}
Mathilde Caron, Hugo Touvron, Ishan Misra, Herv{\'e} J{\'e}gou, Julien Mairal,
  Piotr Bojanowski, and Armand Joulin.
\newblock Emerging properties in self-supervised vision transformers.
\newblock In {\em Proceedings of the IEEE/CVF International Conference on
  Computer Vision}, 2021.

\bibitem{chen2020improved}
Xinlei Chen, Haoqi Fan, Ross Girshick, and Kaiming He.
\newblock Improved baselines with momentum contrastive learning.
\newblock {\em arXiv preprint arXiv:2003.04297}, 2020.

\bibitem{Collins2020EditingIS}
Edo Collins, R. Bala, B. Price, and S. S{\"u}sstrunk.
\newblock Editing in style: Uncovering the local semantics of gans.
\newblock {\em 2020 IEEE/CVF Conference on Computer Vision and Pattern
  Recognition}, 2020.

\bibitem{creswell2018inverting}
Antonia Creswell and Anil~Anthony Bharath.
\newblock Inverting the generator of a generative adversarial network.
\newblock {\em IEEE transactions on neural networks and learning systems},
  2018.

\bibitem{deng2009imagenet}
Jia Deng, Wei Dong, Richard Socher, Li-Jia Li, Kai Li, and Li Fei-Fei.
\newblock Imagenet: A large-scale hierarchical image database.
\newblock In {\em 2009 IEEE conference on computer vision and pattern
  recognition}, 2009.

\bibitem{denton2019detecting}
Emily Denton, Ben Hutchinson, Margaret Mitchell, and Timnit Gebru.
\newblock Detecting bias with generative counterfactual face attribute
  augmentation.
\newblock {\em arXiv preprint arXiv:1906.06439}, 2019.

\bibitem{dinh2022hyperinverter}
Tan~M Dinh, Anh~Tuan Tran, Rang Nguyen, and Binh-Son Hua.
\newblock Hyperinverter: Improving stylegan inversion via hypernetwork.
\newblock In {\em Proceedings of the IEEE/CVF Conference on Computer Vision and
  Pattern Recognition}, 2022.

\bibitem{dosovitskiy2020image}
Alexey Dosovitskiy, Lucas Beyer, Alexander Kolesnikov, Dirk Weissenborn,
  Xiaohua Zhai, Thomas Unterthiner, Mostafa Dehghani, Matthias Minderer, Georg
  Heigold, Sylvain Gelly, et~al.
\newblock An image is worth 16x16 words: Transformers for image recognition at
  scale.
\newblock {\em arXiv preprint arXiv:2010.11929}, 2020.

\bibitem{ge2021revitalizing}
Chongjian Ge, Youwei Liang, Yibing Song, Jianbo Jiao, Jue Wang, and Ping Luo.
\newblock Revitalizing cnn attention via transformers in self-supervised visual
  representation learning.
\newblock In {\em Advances in Neural Information Processing Systems}, 2021.

\bibitem{chongjian-snclr}
Chongjian Ge, Jiangliu Wang, Zhan Tong, Shoufa Chen, Yibing Song, and Ping Luo.
\newblock Soft neighbors are positive supporters in contrastive visual
  representation learning.
\newblock In {\em International Conference on Learning Representations}, 2021.

\bibitem{goetschalckx2019ganalyze}
Lore Goetschalckx, Alex Andonian, Aude Oliva, and Phillip Isola.
\newblock Ganalyze: Toward visual definitions of cognitive image properties.
\newblock In {\em Proceedings of the IEEE/CVF International Conference on
  Computer Vision}, 2019.

\bibitem{grill2020bootstrap}
Jean-Bastien Grill, Florian Strub, Florent Altch{\'e}, Corentin Tallec, Pierre
  Richemond, Elena Buchatskaya, Carl Doersch, Bernardo Avila~Pires, Zhaohan
  Guo, Mohammad Gheshlaghi~Azar, et~al.
\newblock Bootstrap your own latent-a new approach to self-supervised learning.
\newblock {\em Advances in neural information processing systems}, 2020.

\bibitem{gu2020image}
Jinjin Gu, Yujun Shen, and Bolei Zhou.
\newblock Image processing using multi-code gan prior.
\newblock In {\em Proceedings of the IEEE/CVF conference on computer vision and
  pattern recognition}, 2020.

\bibitem{guan2020collaborative}
Shanyan Guan, Ying Tai, Bingbing Ni, Feida Zhu, Feiyue Huang, and Xiaokang
  Yang.
\newblock Collaborative learning for faster stylegan embedding.
\newblock {\em arXiv preprint arXiv:2007.01758}, 2020.

\bibitem{harkonen2020ganspace}
Erik H{\"a}rk{\"o}nen, Aaron Hertzmann, Jaakko Lehtinen, and Sylvain Paris.
\newblock Ganspace: Discovering interpretable gan controls.
\newblock {\em arXiv preprint arXiv:2004.02546}, 2020.

\bibitem{he2020momentum}
Kaiming He, Haoqi Fan, Yuxin Wu, Saining Xie, and Ross Girshick.
\newblock Momentum contrast for unsupervised visual representation learning.
\newblock In {\em Proceedings of the IEEE/CVF conference on computer vision and
  pattern recognition}, 2020.

\bibitem{he2015deep}
Kaiming He, Xiangyu Zhang, Shaoqing Ren, and Jian Sun.
\newblock Deep residual learning for image recognition, 2015.

\bibitem{hou2022feat}
Xianxu Hou, Linlin Shen, Or Patashnik, Daniel Cohen-Or, and Hui Huang.
\newblock Feat: Face editing with attention.
\newblock {\em arXiv preprint arXiv:2202.02713}, 2022.

\bibitem{hu2022style}
Xueqi Hu, Qiusheng Huang, Zhengyi Shi, Siyuan Li, Changxin Gao, Li Sun, and
  Qingli Li.
\newblock Style transformer for image inversion and editing.
\newblock {\em arXiv preprint arXiv:2203.07932}, 2022.

\bibitem{huang2017arbitrary}
Xun Huang and Serge Belongie.
\newblock Arbitrary style transfer in real-time with adaptive instance
  normalization.
\newblock In {\em Proceedings of the IEEE international conference on computer
  vision}, 2017.

\bibitem{huang2020curricularface}
Yuge Huang, Yuhan Wang, Ying Tai, Xiaoming Liu, Pengcheng Shen, Shaoxin Li,
  Jilin Li, and Feiyue Huang.
\newblock Curricularface: adaptive curriculum learning loss for deep face
  recognition.
\newblock In {\em Proceedings of the IEEE/CVF Conference on Computer Vision and
  Pattern Recognition}, 2020.

\bibitem{kang2021gan}
Kyoungkook Kang, Seongtae Kim, and Sunghyun Cho.
\newblock Gan inversion for out-of-range images with geometric transformations,
  2021.

\bibitem{karras2020training}
Tero Karras, Miika Aittala, Janne Hellsten, Samuli Laine, Jaakko Lehtinen, and
  Timo Aila.
\newblock Training generative adversarial networks with limited data.
\newblock {\em Advances in Neural Information Processing Systems},
  33:12104--12114, 2020.

\bibitem{karras2019style}
Tero Karras, Samuli Laine, and Timo Aila.
\newblock A style-based generator architecture for generative adversarial
  networks.
\newblock In {\em Proceedings of the IEEE/CVF conference on computer vision and
  pattern recognition}, pages 4401--4410, 2019.

\bibitem{karras2020analyzing}
Tero Karras, Samuli Laine, Miika Aittala, Janne Hellsten, Jaakko Lehtinen, and
  Timo Aila.
\newblock Analyzing and improving the image quality of stylegan.
\newblock In {\em Proceedings of the IEEE/CVF conference on computer vision and
  pattern recognition}, pages 8110--8119, 2020.

\bibitem{kim2022diffusionclip}
Gwanghyun Kim, Taesung Kwon, and Jong~Chul Ye.
\newblock Diffusionclip: Text-guided diffusion models for robust image
  manipulation.
\newblock In {\em Proceedings of the IEEE/CVF Conference on Computer Vision and
  Pattern Recognition}, 2022.

\bibitem{kim2021exploiting}
Hyunsu Kim, Yunjey Choi, Junho Kim, Sungjoo Yoo, and Youngjung Uh.
\newblock Exploiting spatial dimensions of latent in gan for real-time image
  editing, 2021.

\bibitem{kingma2014adam}
Diederik~P Kingma and Jimmy Ba.
\newblock Adam: A method for stochastic optimization.
\newblock {\em arXiv preprint arXiv:1412.6980}, 2014.

\bibitem{KrauseStarkDengFei-Fei_3DRR2013}
Jonathan Krause, Michael Stark, Jia Deng, and Li Fei-Fei.
\newblock 3d object representations for fine-grained categorization.
\newblock In {\em Proceedings of the IEEE international conference on computer
  vision workshops}, 2013.

\bibitem{lin2021anycost}
Ji Lin, Richard Zhang, Frieder Ganz, Song Han, and Jun-Yan Zhu.
\newblock Anycost gans for interactive image synthesis and editing.
\newblock In {\em Proceedings of the IEEE/CVF Conference on Computer Vision and
  Pattern Recognition}, 2021.

\bibitem{liu2023human}
Hongyu Liu, Xintong Han, ChengBin Jin, Huawei Wei, Zhe Lin, Faqiang Wang, Haoye
  Dong, Yibing Song, Jia Xu, and Qifeng Chen.
\newblock Human motionformer: Transferring human motions with vision
  transformers.
\newblock {\em arXiv preprint arXiv:2302.11306}, 2023.

\bibitem{liu2020rethinking}
Hongyu Liu, Bin Jiang, Yibing Song, Wei Huang, and Chao Yang.
\newblock Rethinking image inpainting via a mutual encoder-decoder with feature
  equalizations.
\newblock In {\em Computer Vision--ECCV 2020: 16th European Conference,
  Glasgow, UK, August 23--28, 2020, Proceedings, Part II 16}, pages 725--741.
  Springer, 2020.

\bibitem{liu2019coherent}
Hongyu Liu, Bin Jiang, Yi Xiao, and Chao Yang.
\newblock Coherent semantic attention for image inpainting.
\newblock In {\em Proceedings of the IEEE/CVF International Conference on
  Computer Vision}, pages 4170--4179, 2019.

\bibitem{liu2021pd}
Hongyu Liu, Ziyu Wan, Wei Huang, Yibing Song, Xintong Han, and Jing Liao.
\newblock Pd-gan: Probabilistic diverse gan for image inpainting.
\newblock In {\em Proceedings of the IEEE/CVF Conference on Computer Vision and
  Pattern Recognition}, pages 9371--9381, 2021.

\bibitem{liu2021deflocnet}
Hongyu Liu, Ziyu Wan, Wei Huang, Yibing Song, Xintong Han, Jing Liao, Bin
  Jiang, and Wei Liu.
\newblock Deflocnet: Deep image editing via flexible low-level controls.
\newblock In {\em Proceedings of the IEEE/CVF Conference on Computer Vision and
  Pattern Recognition}, pages 10765--10774, 2021.

\bibitem{liu2019variance}
Liyuan Liu, Haoming Jiang, Pengcheng He, Weizhu Chen, Xiaodong Liu, Jianfeng
  Gao, and Jiawei Han.
\newblock On the variance of the adaptive learning rate and beyond.
\newblock {\em arXiv preprint arXiv:1908.03265}, 2019.

\bibitem{liu2021swin}
Ze Liu, Yutong Lin, Yue Cao, Han Hu, Yixuan Wei, Zheng Zhang, Stephen Lin, and
  Baining Guo.
\newblock Swin transformer: Hierarchical vision transformer using shifted
  windows.
\newblock In {\em Proceedings of the IEEE/CVF International Conference on
  Computer Vision}, pages 10012--10022, 2021.

\bibitem{liu2015deep}
Ziwei Liu, Ping Luo, Xiaogang Wang, and Xiaoou Tang.
\newblock Deep learning face attributes in the wild, 2015.

\bibitem{luo2017learning}
Junyu Luo, Yong Xu, Chenwei Tang, and Jiancheng Lv.
\newblock Learning inverse mapping by autoencoder based generative adversarial
  nets.
\newblock In {\em International Conference on Neural Information Processing},
  2017.

\bibitem{noble2006support}
William~S Noble.
\newblock What is a support vector machine?
\newblock {\em Nature biotechnology}, 2006.

\bibitem{oord2018representation}
Aaron van~den Oord, Yazhe Li, and Oriol Vinyals.
\newblock Representation learning with contrastive predictive coding.
\newblock {\em arXiv preprint arXiv:1807.03748}, 2018.

\bibitem{pan2021videomoco}
Tian Pan, Yibing Song, Tianyu Yang, Wenhao Jiang, and Wei Liu.
\newblock Videomoco: Contrastive video representation learning with temporally
  adversarial examples.
\newblock In {\em IEEE/CVF Conference on Computer Vision and Pattern
  Recognition}, 2021.

\bibitem{parmar2022spatially}
Gaurav Parmar, Yijun Li, Jingwan Lu, Richard Zhang, Jun-Yan Zhu, and
  Krishna~Kumar Singh.
\newblock Spatially-adaptive multilayer selection for gan inversion and
  editing.
\newblock In {\em Proceedings of the IEEE/CVF Conference on Computer Vision and
  Pattern Recognition}, 2022.

\bibitem{patashnik2021styleclip}
Or Patashnik, Zongze Wu, Eli Shechtman, Daniel Cohen-Or, and Dani Lischinski.
\newblock Styleclip: Text-driven manipulation of stylegan imagery.
\newblock In {\em Proceedings of the IEEE/CVF International Conference on
  Computer Vision}, 2021.

\bibitem{perarnau2016invertible}
Guim Perarnau, Joost van~de Weijer, Bogdan Raducanu, and Jose~M. Álvarez.
\newblock Invertible conditional gans for image editing, 2016.

\bibitem{radford2021learning}
Alec Radford, Jong~Wook Kim, Chris Hallacy, Aditya Ramesh, Gabriel Goh,
  Sandhini Agarwal, Girish Sastry, Amanda Askell, Pamela Mishkin, Jack Clark,
  et~al.
\newblock Learning transferable visual models from natural language
  supervision.
\newblock In {\em International Conference on Machine Learning}, 2021.

\bibitem{richardson2020encoding}
Elad Richardson, Yuval Alaluf, Or Patashnik, Yotam Nitzan, Yaniv Azar, Stav
  Shapiro, and Daniel Cohen-Or.
\newblock Encoding in style: a stylegan encoder for image-to-image translation.
\newblock In {\em Proceedings of the IEEE/CVF Conference on Computer Vision and
  Pattern Recognition}, 2021.

\bibitem{roich2021pivotal}
Daniel Roich, Ron Mokady, Amit~H Bermano, and Daniel Cohen-Or.
\newblock Pivotal tuning for latent-based editing of real images.
\newblock {\em arXiv preprint arXiv:2106.05744}, 2021.

\bibitem{shen2020interpreting}
Yujun Shen, Jinjin Gu, Xiaoou Tang, and Bolei Zhou.
\newblock Interpreting the latent space of gans for semantic face editing.
\newblock In {\em Proceedings of the IEEE/CVF Conference on Computer Vision and
  Pattern Recognition}, 2020.

\bibitem{shen2020closedform}
Yujun Shen and Bolei Zhou.
\newblock Closed-form factorization of latent semantics in gans.
\newblock {\em arXiv preprint arXiv:2007.06600}, 2020.

\bibitem{tov2021designing}
Omer Tov, Yuval Alaluf, Yotam Nitzan, Or Patashnik, and Daniel Cohen-Or.
\newblock Designing an encoder for stylegan image manipulation, 2021.

\bibitem{vaswani2017attention}
Ashish Vaswani, Noam Shazeer, Niki Parmar, Jakob Uszkoreit, Llion Jones,
  Aidan~N Gomez, {\L}ukasz Kaiser, and Illia Polosukhin.
\newblock Attention is all you need.
\newblock {\em Advances in neural information processing systems}, 2017.

\bibitem{voynov2020unsupervised}
Andrey Voynov and Artem Babenko.
\newblock Unsupervised discovery of interpretable directions in the gan latent
  space.
\newblock {\em arXiv preprint arXiv:2002.03754}, 2020.

\bibitem{wang2022high}
Tengfei Wang, Yong Zhang, Yanbo Fan, Jue Wang, and Qifeng Chen.
\newblock High-fidelity gan inversion for image attribute editing.
\newblock In {\em Proceedings of the IEEE/CVF Conference on Computer Vision and
  Pattern Recognition}, 2022.

\bibitem{wei2021hairclip}
Tianyi Wei, Dongdong Chen, Wenbo Zhou, Jing Liao, Zhentao Tan, Lu Yuan, Weiming
  Zhang, and Nenghai Yu.
\newblock Hairclip: Design your hair by text and reference image.
\newblock {\em arXiv preprint arXiv:2112.05142}, 2021.

\bibitem{xia2021tedigan}
Weihao Xia, Yujiu Yang, Jing-Hao Xue, and Baoyuan Wu.
\newblock Tedigan: Text-guided diverse face image generation and manipulation.
\newblock In {\em IEEE Conference on Computer Vision and Pattern Recognition},
  2021.

\bibitem{xia2021gan}
Weihao Xia, Yulun Zhang, Yujiu Yang, Jing-Hao Xue, Bolei Zhou, and Ming-Hsuan
  Yang.
\newblock Gan inversion: A survey, 2021.

\bibitem{xuyao2022}
Xu Yao, Alasdair Newson, Yann Gousseau, and Pierre Hellier.
\newblock A style-based gan encoder for high fidelity reconstruction of images
  and videos.
\newblock {\em European conference on computer vision}, 2022.

\bibitem{yeh2017semantic}
Raymond~A. Yeh, Chen Chen, Teck~Yian Lim, Alexander~G. Schwing, Mark
  Hasegawa-Johnson, and Minh~N. Do.
\newblock Semantic image inpainting with deep generative models, 2017.

\bibitem{yu2016lsun}
Fisher Yu, Ari Seff, Yinda Zhang, Shuran Song, Thomas Funkhouser, and Jianxiong
  Xiao.
\newblock Lsun: Construction of a large-scale image dataset using deep learning
  with humans in the loop, 2016.

\bibitem{yuksel2021latentclr}
O{\u{g}}uz~Kaan Y{\"u}ksel, Enis Simsar, Ezgi~G{\"u}lperi Er, and Pinar
  Yanardag.
\newblock Latentclr: A contrastive learning approach for unsupervised discovery
  of interpretable directions.
\newblock In {\em Proceedings of the IEEE/CVF International Conference on
  Computer Vision}, 2021.

\bibitem{zhang2019lookahead}
Michael Zhang, James Lucas, Jimmy Ba, and Geoffrey~E Hinton.
\newblock Lookahead optimizer: k steps forward, 1 step back.
\newblock {\em Advances in neural information processing systems}, 2019.

\bibitem{zhang2018perceptual}
Richard Zhang, Phillip Isola, Alexei~A Efros, Eli Shechtman, and Oliver Wang.
\newblock The unreasonable effectiveness of deep features as a perceptual
  metric.
\newblock In {\em Proceedings of the IEEE conference on computer vision and
  pattern recognition}, 2018.

\bibitem{zhang2020contrastive}
Yuhao Zhang, Hang Jiang, Yasuhide Miura, Christopher~D Manning, and Curtis~P
  Langlotz.
\newblock Contrastive learning of medical visual representations from paired
  images and text.
\newblock {\em arXiv preprint arXiv:2010.00747}, 2020.

\bibitem{zhu2020domain}
Jiapeng Zhu, Yujun Shen, Deli Zhao, and Bolei Zhou.
\newblock In-domain gan inversion for real image editing.
\newblock {\em arXiv preprint arXiv:2004.00049}, 2020.

\bibitem{zhu2016generative}
Jun-Yan Zhu, Philipp Kr{\"a}henb{\"u}hl, Eli Shechtman, and Alexei~A Efros.
\newblock Generative visual manipulation on the natural image manifold.
\newblock In {\em European conference on computer vision}. Springer, 2016.

\bibitem{zhu2020improved}
Peihao Zhu, Rameen Abdal, Yipeng Qin, and Peter Wonka.
\newblock Improved stylegan embedding: Where are the good latents?, 2020.

\bibitem{zhu2022one}
Yiming Zhu, Hongyu Liu, Yibing Song, Xintong Han, Chun Yuan, Qifeng Chen, Jue
  Wang, et~al.
\newblock One model to edit them all: Free-form text-driven image manipulation
  with semantic modulations.
\newblock {\em arXiv preprint arXiv:2210.07883}, 2022.

\end{thebibliography}
}
\renewcommand\thesection{\Alph{section}}
\setcounter{section}{0}
\newpage

In this supplementary material, we first describe the limitation of our method. Then, we present more analysis about our solid foundation latent code $w$. Meanwhile, we show more visual comparisons of the ClebA-HQ~\cite{liu2015deep} and car datasets~\cite{yu2016lsun}. Finally, we demonstrate our method can achieve good performance in the horse dataset~\cite{yu2016lsun} in visually.
\section{Limitation}
Our method has good performance in both qualitative and quantitative, but it still has some limitations. Our method cannot   reconstruct the jewelry well of some corner cases, and there are some artifacts during the editing process. We can replace the CNN with a more powerful network(i.e., Vision Transformer~\cite{dosovitskiy2020image,liu2021swin}) to try to solve these problems.

\begin{figure*}[t]
\begin{center}
\includegraphics[width=0.98\linewidth]{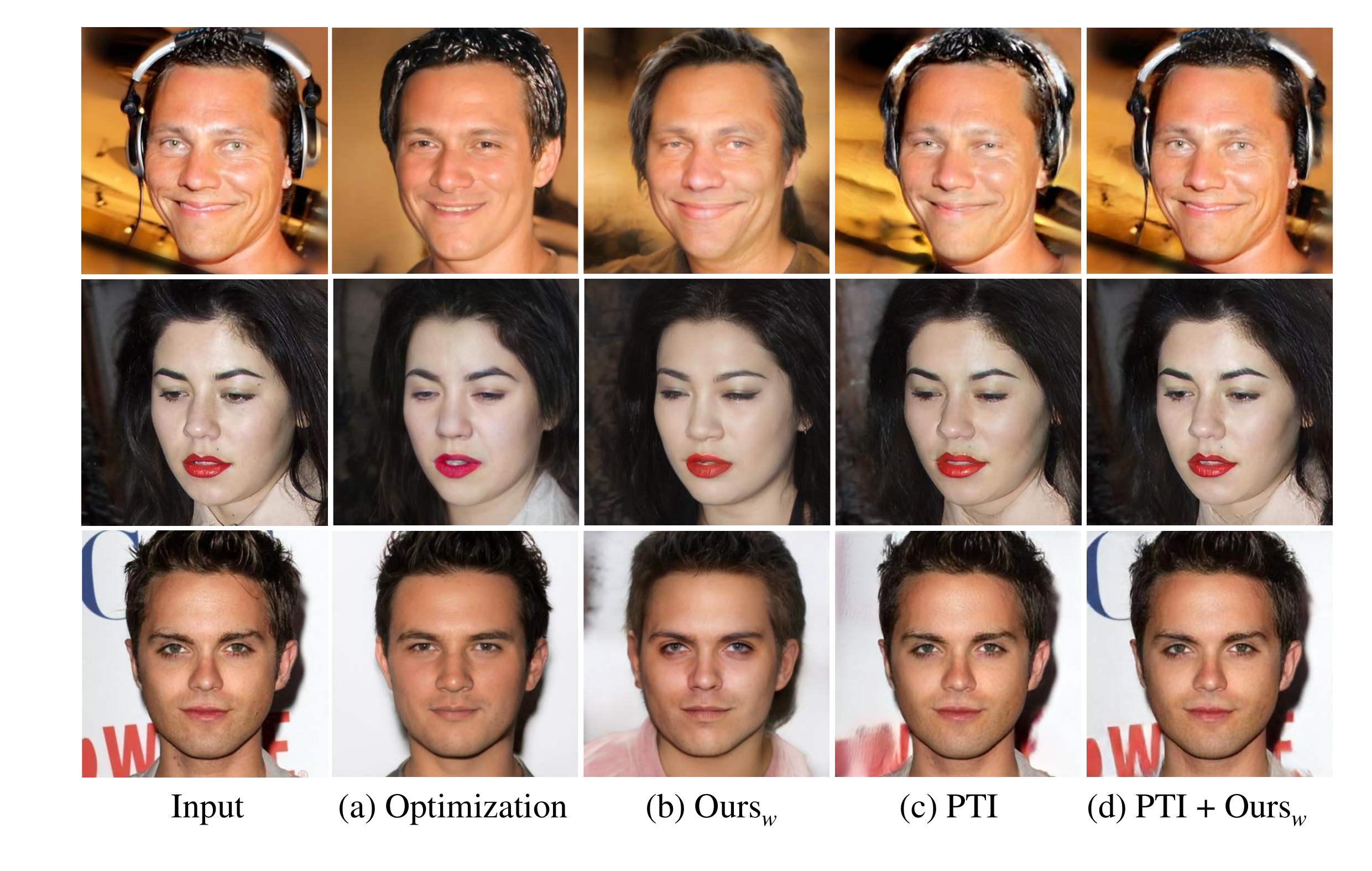}
\caption{Analysis of latent code $w$. We replace the initialization of PTI with our $w$ as shown in (d). The original PTI's result is (c). We can find that our solid latent code $w$ can help the PTI perform better. Meanwhile, we illustrate the reconstruction results with optimization $w$ and our $w$ in (a) and (b), respectively.  }
\vspace{-0.1in}
\label{fig:PTI+oursW}
\end{center}
\end{figure*}

\section{More Analysis}
\begin{figure*}[t]
\begin{center}
\includegraphics[width=1\linewidth]{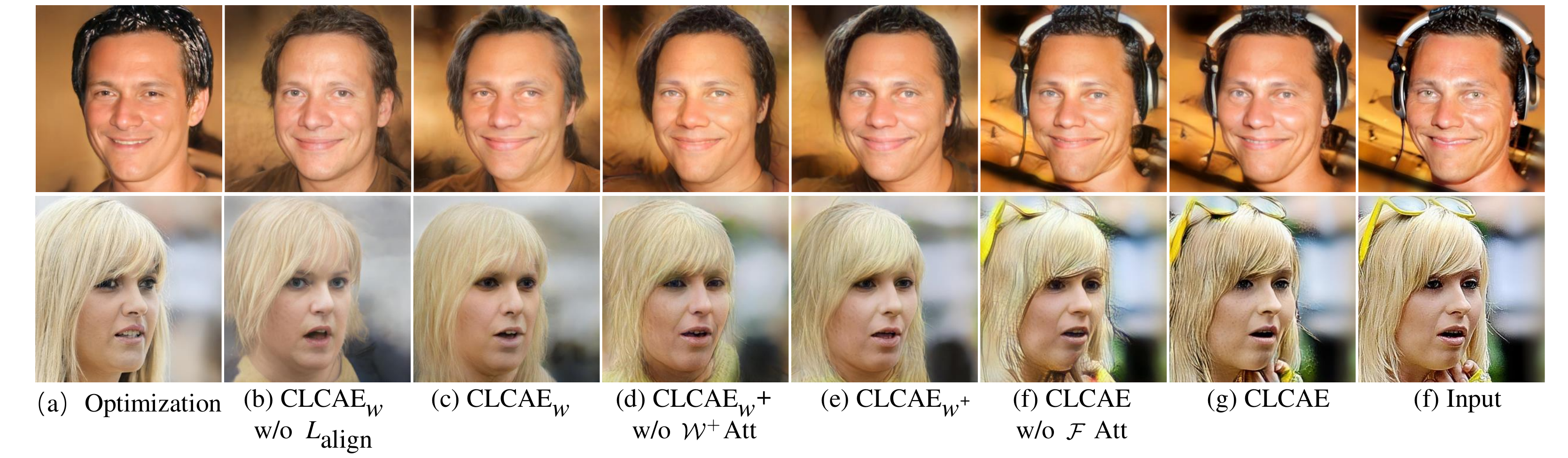}

\caption{Qualitative ablation}
\vspace{-0.35in}
\label{fig:moreablation}
\end{center}
\end{figure*}

To further prove that our method can predict robust latent code $w$. We set our $w$ as the initialization of PTI~\cite{roich2021pivotal} to make comparisons. As shown in Fig.~\ref{fig:PTI+oursW}, the (a) is the original initialization results with $w$ in PTI,  and PTI finds this $w$ with the optimization method~\cite{karras2020analyzing}. The (b) is the reconstruction results with our $w$, and the (b) outperformance than (a) in both identity and detail preservation which verifies the effectiveness of our method. The (c) is the original final prediction of PTI which sets the optimization   $w$  as the initialization, and we replace the optimization $w$ with our  $w$ to get (d). By comparing (c)  and (d), we can find a robust $w$ that can improve the performance of PTI. Meanwhile, since the $w$ in (d) is predicted with our encoder, we can speed PTI up to 134s  for a single image, which is almost half of the time-consuming of the original PTI. Moreover, we provide more visual results of ablation study~\ref{fig:moreablation}.

\section{More visual comparisons}

\begin{figure*}[t]
\begin{center}
\includegraphics[width=1\linewidth]{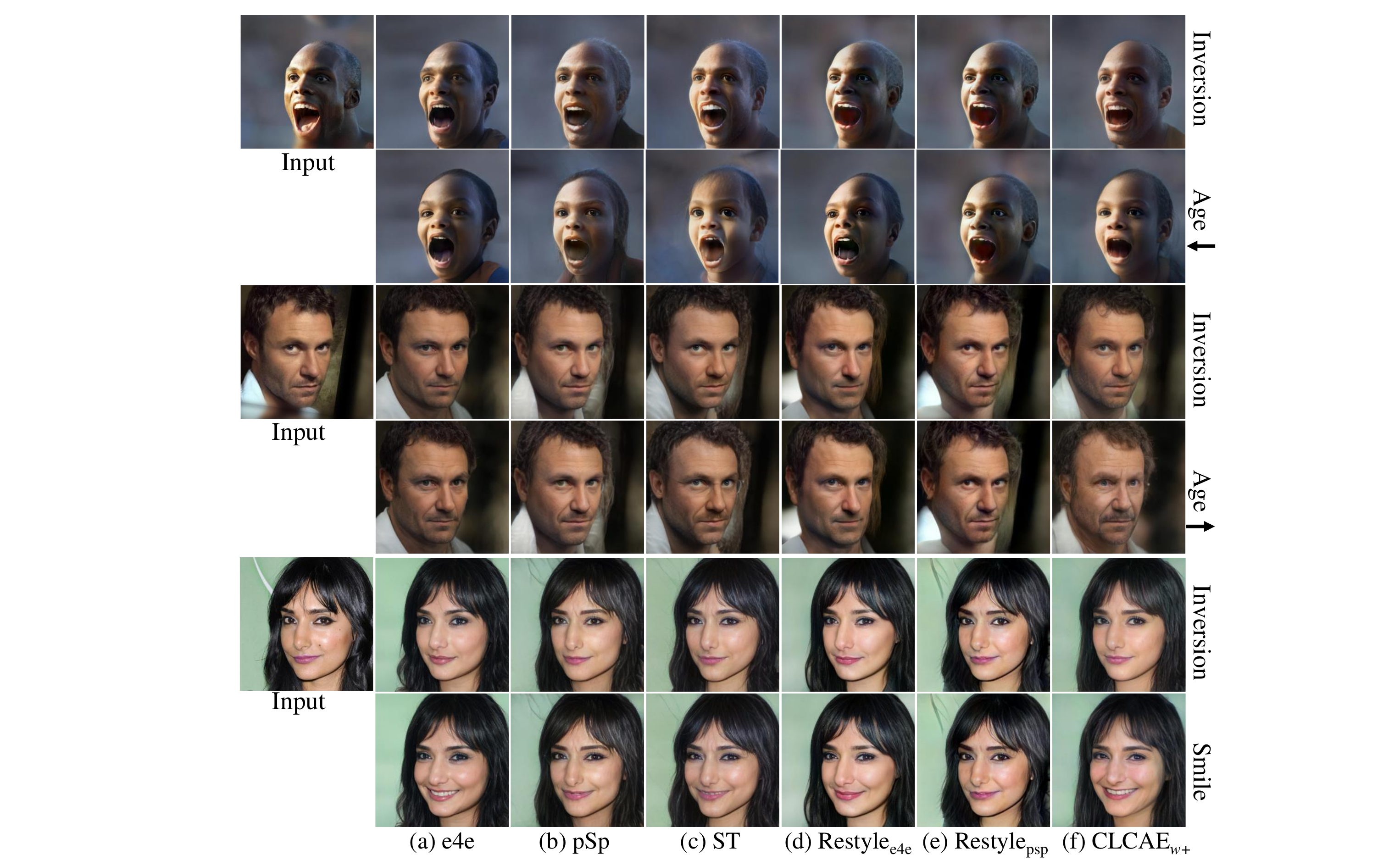}
\caption{More visual comparisons on ClebA-HQ~\cite{liu2015deep} dataset for $\mathcal{W^+}$ space methods. Our method performance better in both reconstruction and editing. $\downarrow$ means a reduction of the manipulation attribute. $\uparrow$ means an increment of the manipulation attribute. }
\vspace{-0.1in}
\label{fig:W+_supp_face}
\end{center}
\end{figure*}

\begin{figure*}[t]
\begin{center}
\includegraphics[width=1\linewidth]{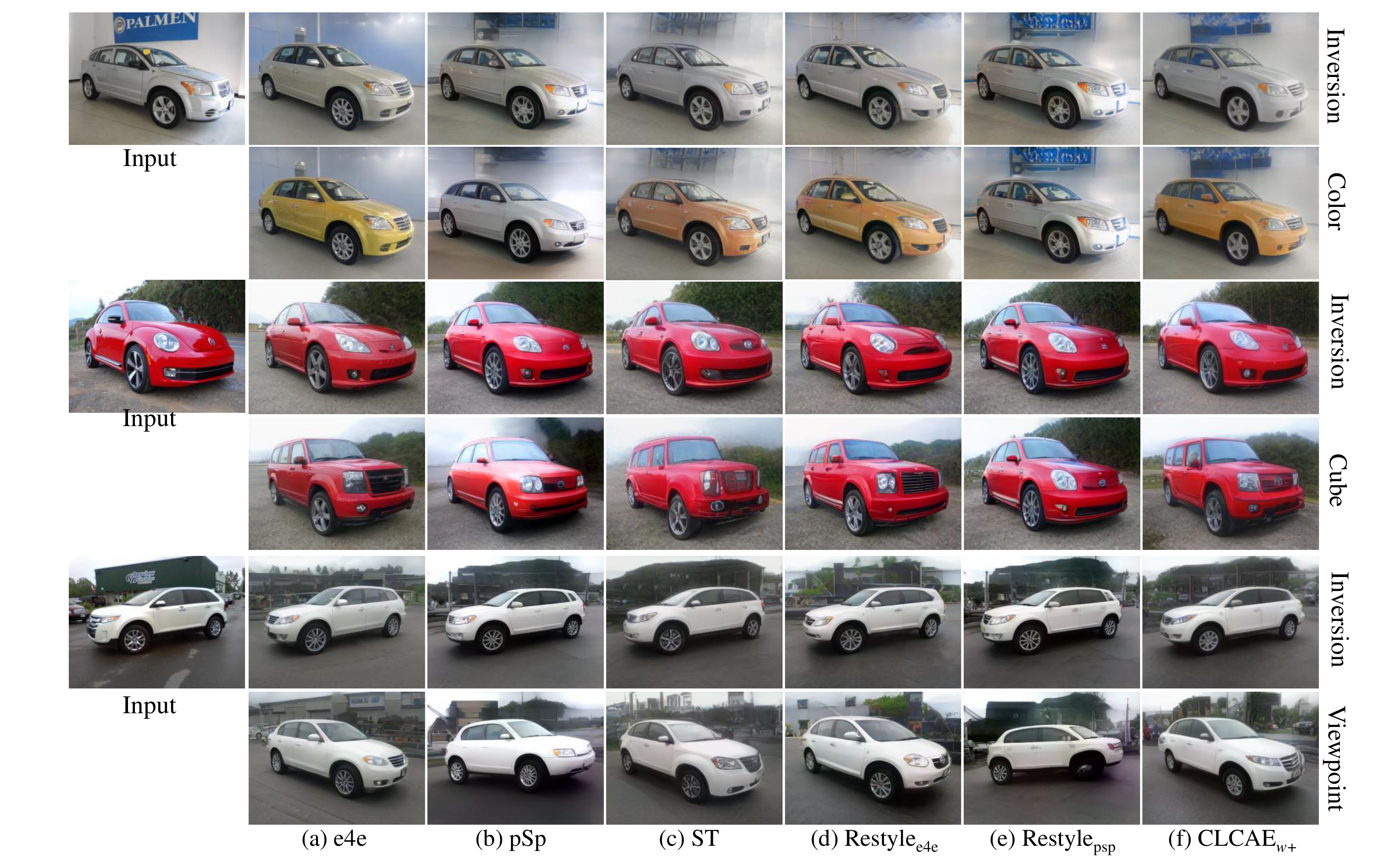}
\caption{More visual comparisons on car dataset~\cite{yu2016lsun} for $\mathcal{W^+}$ space methods. Our method performance better in both reconstruction and editing.  }
\vspace{-0.1in}
\label{fig:W+_supp_car}
\end{center}
\end{figure*}
{\flushleft \bf $\mathcal{W^+}$ space.} We show more visual comparisons between $\mathcal{W^+}$ space methods (e4e~\cite{tov2021designing}, pSp~\cite{richardson2020encoding}, restyle$_{\text{pSp}}$~\cite{alaluf2021restyle}, restyle$_{\text{e4e}}$~\cite{alaluf2021restyle} and  StyleTransformer (ST)~\cite{hu2022style})  and our method in  Fig.~\ref{fig:W+_supp_face} and  Fig.~\ref{fig:W+_supp_car}. Except for the e4e and our method, the other methods seem to have an overfitting phenomenon (i.e., the wrong white hair in the (c),(d), and (e) of the second person in Fig.~\ref{fig:W+_supp_face}) as discussed in our main paper. Meanwhile, our method has better reconstruction and editing performance simultaneously than other baselines (i.e., the "Age" and "Smile" editing results in Fig.~\ref{fig:W+_supp_face} and the "Viewpoint"  editing results in Fig.~\ref{fig:W+_supp_car}).

\begin{figure*}[t]
\begin{center}
\includegraphics[width=1\linewidth]{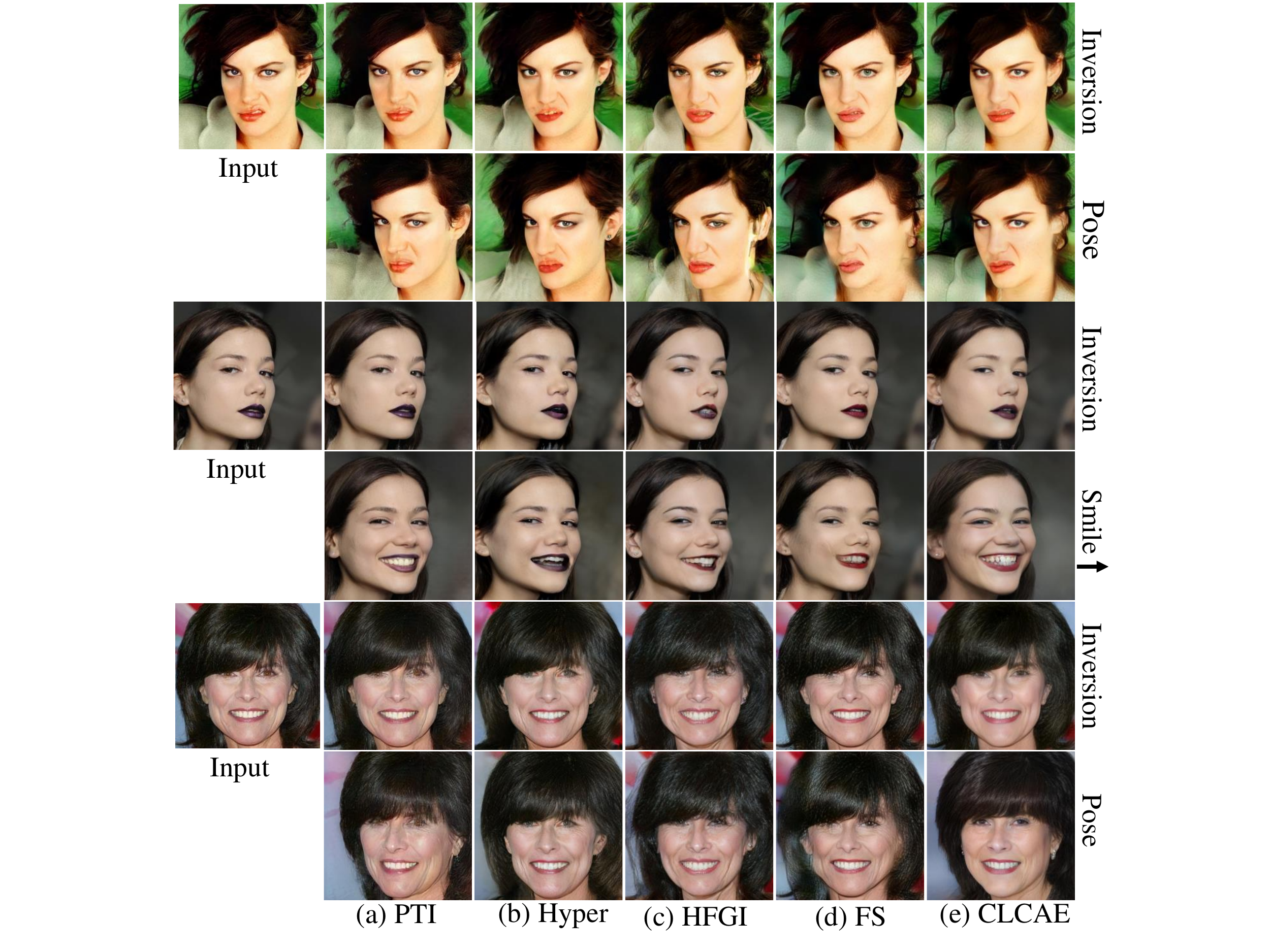}
\caption{More visual comparisons on ClebA-HQ~\cite{liu2015deep} dataset for $\mathcal{F}$ space methods. Our method performance better in both reconstruction and editing.  $\uparrow$ means an increment of the manipulation attribute.}
\vspace{-0.1in}
\label{fig:F_supp_Face}
\end{center}
\end{figure*}

\begin{figure*}[t]
\begin{center}
\includegraphics[width=0.98\linewidth]{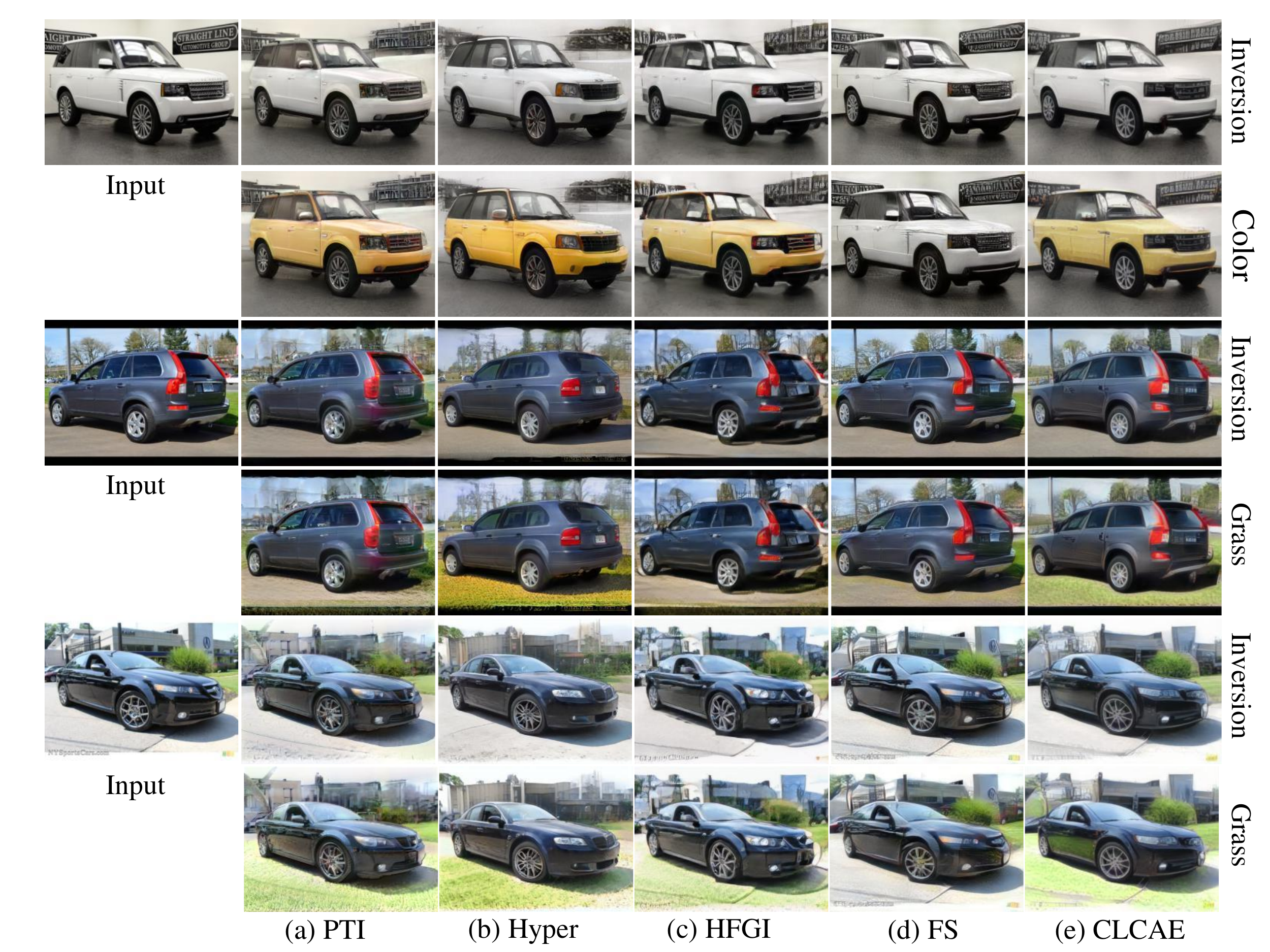}
\caption{More visual comparisons on car dataset~\cite{yu2016lsun} for $\mathcal{F}$ space methods. Our method performance better in both reconstruction and editing.   }
\vspace{-0.1in}
\label{fig:F_supp_Car}
\end{center}
\end{figure*}

\begin{figure*}[t]
\begin{center}
\includegraphics[width=1\linewidth]{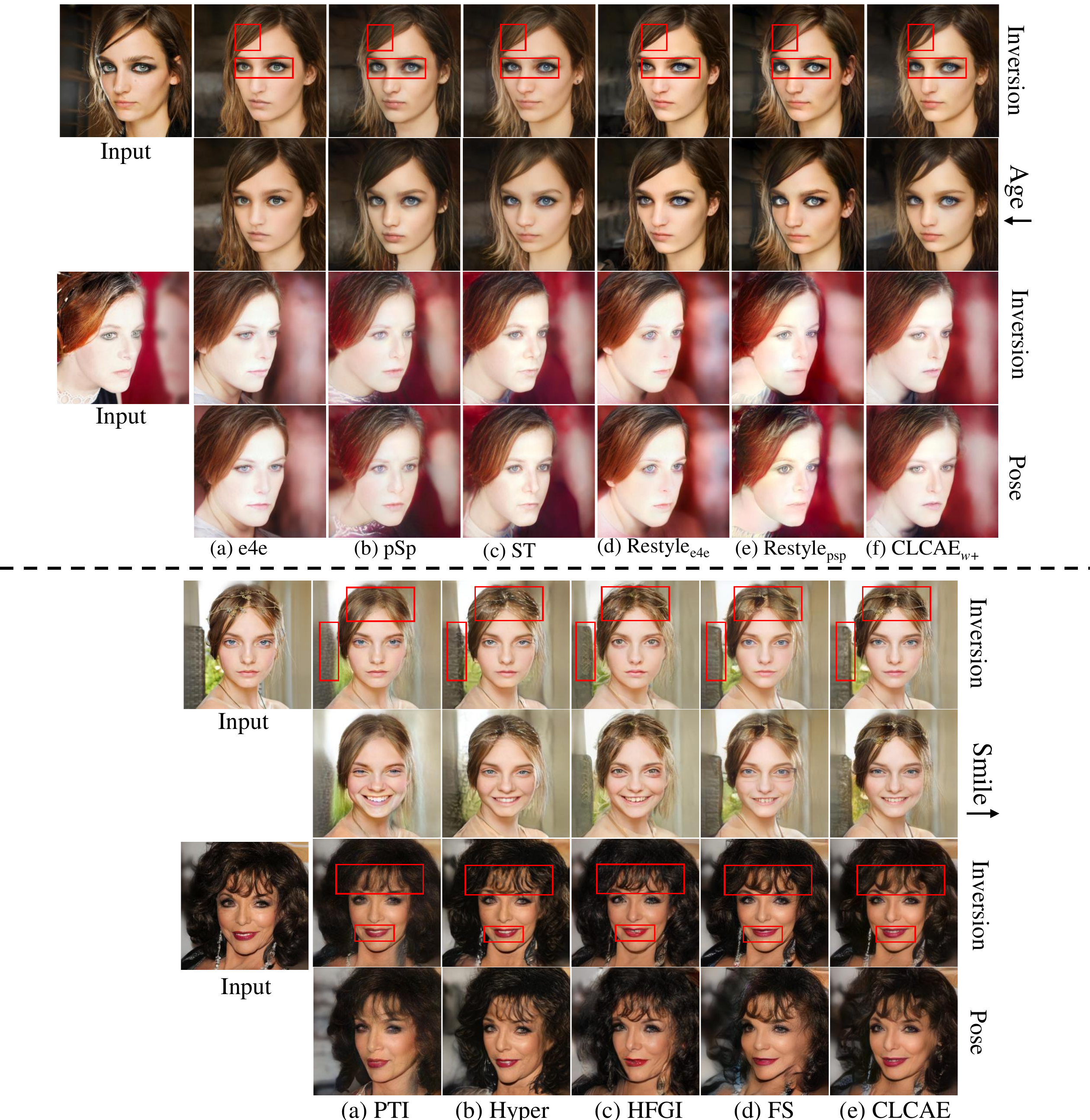}

\caption{More visual comparisons.}
\vspace{-0.35in}
\label{fig:all}
\end{center}
\end{figure*}

{\flushleft \bf $\mathcal{F}$ space.}  Fig.~\ref{fig:F_supp_Face} and Fig.~\ref{fig:F_supp_Car} shows more our comparisons to PTI~\cite{roich2021pivotal}, Hyper~\cite{alaluf2022hyperstyle}, HFGI~\cite{wang2022high}, and FS~\cite{xuyao2022}  in the $\mathcal{F}$ space. Our method can produce the image with better quality in both reconstruction and editing than other baselines (i.e., the "Pose"  editing results in Fig.~\ref{fig:F_supp_Face}  and the "Grass"  editing results in Fig.~\ref{fig:F_supp_Car}). 

Moreover, we show more visual comparisons in Fig.~\ref{fig:all}.

\section{More visual results}
\begin{figure*}[t]
\begin{center}
\includegraphics[width=0.98\linewidth]{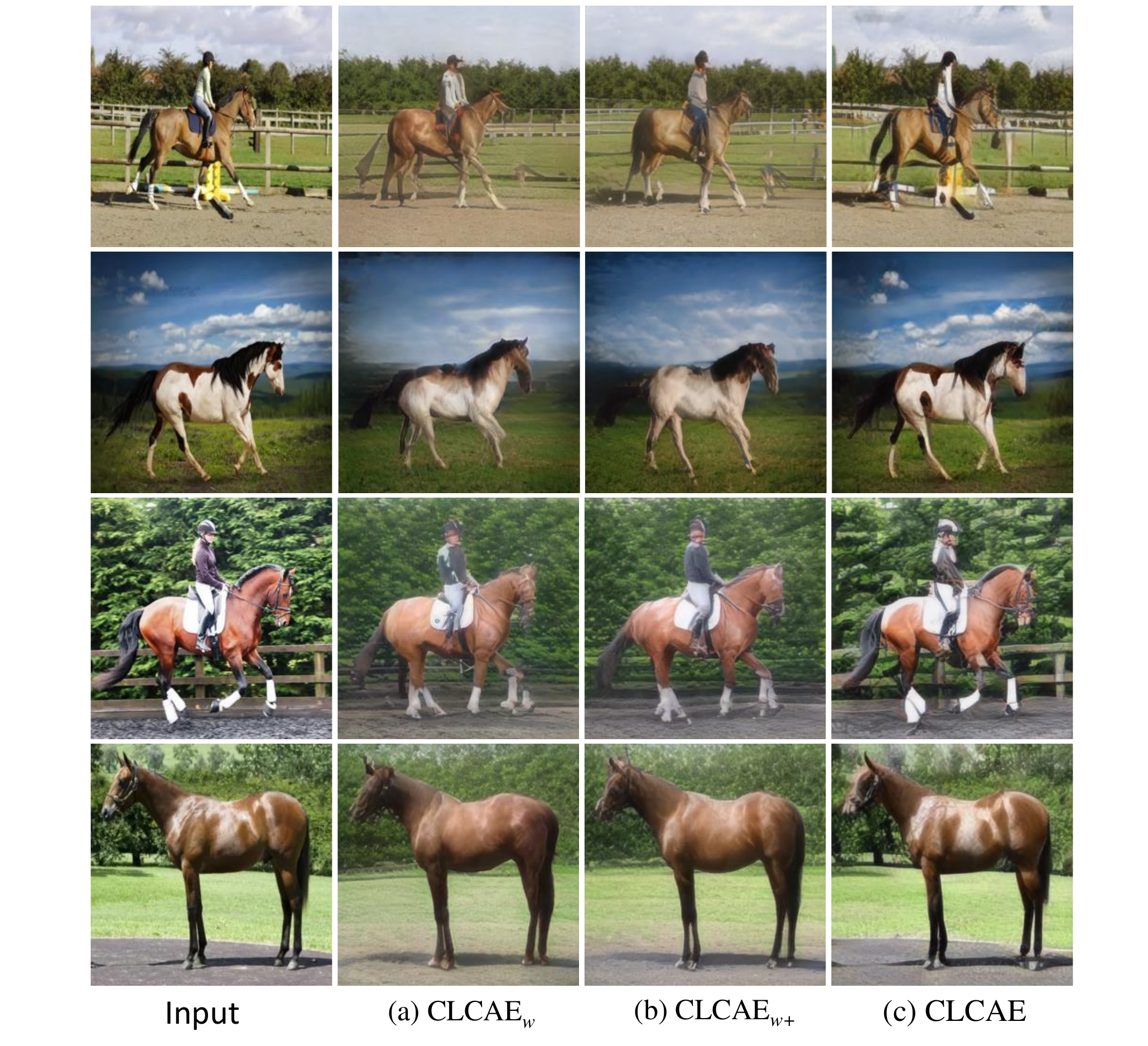}
\caption{More visual results on horse dataset~\cite{yu2016lsun}. Good results can demonstrate the robustness of our method.   }
\vspace{-0.1in}
\label{fig:horse}
\end{center}
\end{figure*}

In addition to the face and car datasets, we also show more visual results on horse dataset~\cite{yu2016lsun} in Fig~\ref{fig:horse}. We show the reconstruction results with our $w$, $w^+$ and $w^+ , f$ in (a) , (b) and (c) respectively. These visual results show that our solid foundation latent code $w$ method can produce good-quality reconstruction images, and our $w^+$ and $f$ can further generate high-fidelity results with the solid $w$.

\end{document}